\documentclass[10pt,twocolumn,letterpaper]{article}

\usepackage[pagenumbers]{config/cvpr} %

\usepackage{tikz}
\usepackage{enumitem}
\usepackage{multirow}
\usepackage{adjustbox}
\usepackage{amsmath}

\usepackage{makecell}
\usepackage{pifont}%

\definecolor{cvprblue}{rgb}{0.21,0.49,0.74}
\usepackage[pagebackref,breaklinks,colorlinks,citecolor=cvprblue]{hyperref}

\usepackage{fontawesome}

\usepackage{lipsum}

\usepackage{graphicx}
\usepackage{amsmath}
\usepackage{amssymb}
\usepackage{booktabs}
\usepackage{blindtext}

\definecolor{citecolor}{HTML}{0071bc}
\definecolor{frontcolor}{HTML}{325ea5}
\definecolor{sidecolor}{HTML}{a58b77}
\definecolor{DeltaColor}{rgb}{0.039,0.73,0.71}
\definecolor{SigmaColor}{rgb}{0.98,0.45,0.0}
\definecolor{AlphaColor}{rgb}{0,0,0.8}
\definecolor{BetaColor}{rgb}{0.8,0,0.8}
\definecolor{GammaColor}{rgb}{0.514,0.34,0.224}
\definecolor{EpsilonColor}{rgb}{0.353,0.725,0.906}
\definecolor{PurpleColor}{HTML}{bca5ea}
\definecolor{OrangeColor}{rgb}{0.914,0.541,0.0.141}
\definecolor{GreenColor}{rgb}{0.137,0.573,0.565}
\definecolor{RedColor}{rgb}{0.949,0.275, 0.224}
\definecolor{LightCyan}{rgb}{0.88,1,1}
\definecolor{Gray}{gray}{0.85}
\definecolor{LilacColor}{HTML}{8D538D} %

\newcommand{\orange}[1]{\textcolor{OrangeColor}{#1}}

\newcommand{\lilac}[1]{\textcolor{LilacColor}{#1}}

\newcommand{\highlight}[1]{{\color{black}#1}}

\newcommand{\nameMethod}{\mbox{InteractVLM}\xspace}

\newcommand{\nameRepresentation}{\mbox{RLL}\xspace}
\newcommand{\nameRepresentationLONG}{\mbox{Render-Localize-Lift}\xspace}

\newcommand{\Render}{{Render}\xspace}

\newcommand{\Lift}{{Lift}\xspace}

\newcommand{\smpl}{\mbox{SMPL}\xspace}
\newcommand{\smplx}{\mbox{SMPL-X}\xspace}

\newcommand{\smplX}{\smplx}
\newcommand{\smplh}{\mbox{SMPL+H}\xspace}
\newcommand{\smplH}{\smplh}

\newcommand{\SAM}{\mbox{SAM}\xspace}
\newcommand{\PIAD}{\mbox{PIAD}\xspace}
\newcommand{\LEMON}{\mbox{LEMON}\xspace}
\newcommand{\CHORE}{\mbox{CHORE}\xspace}
\newcommand{\CONTHO}{\mbox{CONTHO}\xspace}
\newcommand{\HDM}{\mbox{HDM}\xspace}

\newcommand{\GPT}{\mbox{GPT4o}\xspace}
\newcommand{\VQA}{\mbox{VQA}\xspace}

\newcommand{\LLAVA}{\mbox{LLaVA}\xspace}
\newcommand{\LORA}{\mbox{LoRA}\xspace}

\newcommand{\OpenShape}{\mbox{OpenShape}\xspace}
\newcommand{\VLM}{\mbox{VLM}\xspace}
\newcommand{\VLMs}{\mbox{VLMs}\xspace}
\newcommand{\MVLoc}{\mbox{MV-Loc}\xspace}
\newcommand{\HOI}{\mbox{HOI}\xspace}
\newcommand{\semanticDECO}{Semantic-DECO\xspace}

\newcommand{\FeatLift}{\mbox{FeatLift}\xspace}

\usepackage[normalem]{ulem}
\usepackage{lipsum}
\newcommand{\BSTRO}{\mbox{BSTRO}\xspace}
\newcommand{\DECO}{\mbox{DECO}\xspace}
\newcommand{\DAMON}{\mbox{DAMON}\xspace}
\newcommand{\PHOSA}{\mbox{PHOSA}\xspace}

\newcommand{\POSA}{\mbox{POSA}\xspace}
\newcommand{\PIXIE}{\mbox{PIXIE}\xspace}
\newcommand{\OSX}{\mbox{OSX}\xspace}

\newcommand{\sota}{{state-of-the-art}\xspace}
\newcommand{\SOTA}{\mbox{SotA}\xspace}

\newcommand{\inthewild}{{in-the-wild}\xspace}

\newcommand{\zheading}[1]{\textbf{#1.}}
\newcommand{\qheading}[1]{\noindent\textbf{#1.}}

\renewcommand{\etal}{\mbox{et al.}\xspace}
\renewcommand{\ie}{\mbox{i.e.}\xspace}
\renewcommand{\eg}{\mbox{e.g.}\xspace}
\renewcommand{\wrt}{\mbox{w.r.t.}\xspace}

\newcommand{\CR}[1]{{\color{black} #1}}

\newcommand{\moveToSupMat}[1]{\begin{comment}#1\end{commment}}
\newcommand{\supmat}{{{Sup.~Mat}}.\xspace}

\usepackage{textcomp}

\usepackage{multirow}

\usepackage{txfonts}

\usepackage{tikz}
\usepackage{overpic}

\usepackage{pifont}
\newcommand{\cmark}{\color{ForestGreen}\ding{51}}

\newcommand{\xmark}{\color{red}\ding{55}}

\newcommand{\image}{I}

\newcommand{\inputPrompt}{T_{inp}}
\newcommand{\outputPrompt}{T_{out}}
\newcommand{\humanToken}{{{\textless{\tt HCON}\textgreater}}\xspace}
\newcommand{\objectToken}{{{\textless{\tt OCON}\textgreater}}\xspace}
\newcommand{\objectEmd}{E^{O}}
\newcommand{\humanEmd}{E^{H}}
\newcommand{\humanObjectEmd}{E^{H,O}}

\newcommand{\humanObjectEmdLift}{E^{H,O}_{3D}}

\newcommand{\jviews}{J}
\newcommand{\camParams}{{K}}
\newcommand{\inputRender}{R}
\newcommand{\inputRenderOH}{R^{{{H,O}}}}
\newcommand{\mask}{M}
\newcommand{\validmask}{\mask}
\newcommand{\maskP}{p_M}

\newcommand{\validmaskP}{p_{\validmask}}
\newcommand{\validmaskGT}{\widehat{\mask}}
\newcommand{\humanMask}{\mask^H}
\newcommand{\objectMask}{\mask^O}
\newcommand{\humanObjectMask}{\mask^{H,O}}

\newcommand{\vlmNN}{\Psi}
\newcommand{\imageEncoder}{\Theta}
\newcommand{\HmaskDecoder}{\Omega^H}
\newcommand{\OmaskDecoder}{\Omega^O}
\newcommand{\HOmaskDecoder}{\Omega^{H,O}}
\newcommand{\featLiftNN}{\Phi}
\newcommand{\projLayerNN}{\Gamma}

\newcommand{\tokenLoss}{\mathcal{L}_{token}}
\newcommand{\bceLoss}{\mathcal{L}_{BCE}}
\newcommand{\diceLoss}{\mathcal{L}_{Dice}}
\newcommand{\hCLoss}{\mathcal{L}_{C}^{H}}
\newcommand{\oCLoss}{\mathcal{L}_{C}^{O}}

\newcommand{\human}{{H}}
\newcommand{\object}{{O}}
\newcommand{\vertices}{V}

\newcommand{\hcontact}{C^{H}}
\newcommand{\ocontact}{C^{O}}
\newcommand{\ocontactGT}{\widehat{C}^{O}}
\newcommand{\hcontactP}{p_{hC}}
\newcommand{\ohcontact}{C^{H,O}}

\newcommand{\humanMesh}{\mathcal{H}}

\newcommand{\objectMesh}{\mathcal{O}}

\newcommand{\objectRot}{R^{O}}
\newcommand{\objectTransl}{t^{O}}
\newcommand{\objectScale}{s^{O}}

\newcommand{\gtMask}{M}
\newcommand{\gtMeanMaskPxl}{\gtMask^c}

\newcommand{\rendMask}{\widehat{\gtMask}}
\newcommand{\rendMeanMaskPxl}{\widehat{\gtMask^c}}

\newcommand{\rendDepth}{\widehat{D}}

\newcommand{\LISA}{\mbox{LISA}}

\title{\vspace{-0.7 em}\nameMethod: 3D Interaction Reasoning from 2D Foundational Models\vspace{-0.35 em}}

\author{
    Sai Kumar Dwivedi$^{1}$ \quad 
    Dimitrije Anti\'{c}$^{2}$    \quad 
    Shashank Tripathi$^{1}$ \quad
    Omid Taheri$^{1}$ \quad \\
    Cordelia Schmid$^{3}$\quad
    Michael J. Black$^{1}$ \quad
    Dimitrios Tzionas$^{2}$ \quad 
    \vspace{0.7em}\\
    \normalsize 
    $^1$Max Planck Institute for Intelligent Systems, T\"{u}bingen, Germany \quad
    \normalsize 
    $^2$University of Amsterdam (UvA), the Netherlands \\
    \normalsize 
    $^3$Inria, \'Ecole normale sup\'erieure, CNRS, PSL Research University, France\\
}

\begin{document}
\twocolumn[{%
\renewcommand\twocolumn[1][]{#1}%
\maketitle
 \vspace*{-2.0 em}
\begin{center}
    \centering
    \captionsetup{type=figure}%
        \includegraphics[width=0.99 \linewidth]{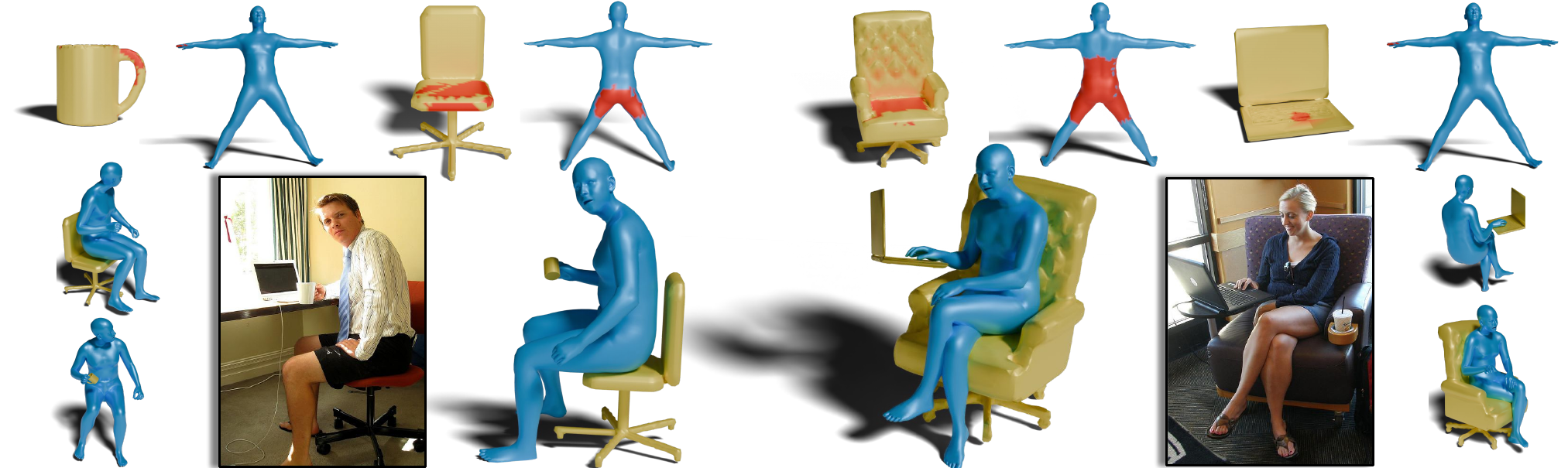}
    \vspace{-0.5 em}
    \captionof{figure}{
        We present \nameMethod, a novel method for estimating contact points on both human bodies and objects from a single \inthewild image, 
        shown here 
        as red patches. 
        Our method goes beyond traditional binary contact estimation methods by estimating contact points on a human in relation to a specified object.
        We do so by leveraging the broad visual knowledge of a large Visual Language Model.
    }
    \vspace{+0.5 em}
\label{fig:teaser}
\end{center}%
}]

\begin{abstract} 

We introduce \nameMethod, a novel method to estimate 3D contact points on human bodies and objects from single in-the-wild images, enabling accurate human-object joint reconstruction in 3D.
This is challenging due to occlusions, depth ambiguities, and widely varying object shapes.
Existing methods rely on 3D contact annotations collected via expensive motion-capture systems or tedious manual labeling, limiting scalability and generalization.
To overcome this, \nameMethod harnesses the broad visual knowledge of large Vision-Language Models (\VLMs), fine-tuned with limited 3D contact data.
However, directly applying these models
is non-trivial, as they reason ``only'' in 2D, while human-object contact is inherently 3D.
Thus we introduce a novel “Render-Localize-Lift” module that: (1) embeds 3D body and object surfaces in 2D space via multi-view rendering, (2) trains a novel multi-view localization model (MV-Loc) to infer contacts in 2D, and (3) lifts these to 3D.
Additionally, we propose a new task called \textit{Semantic Human Contact} estimation, where human contact predictions are conditioned explicitly on object semantics, enabling richer interaction modeling.
\nameMethod outperforms existing work on contact estimation and also facilitates 3D reconstruction from an in-the-wild image. 
To estimate 3D human and object pose, we infer initial body
and object meshes, then infer contacts on both of these via
\nameMethod, and last exploit these for fitting the meshes to
image evidence. Results show that our approach performs
promisingly in the wild.
Code and models are available at 
\url{https://interactvlm.is.tue.mpg.de}. 
\end{abstract}
    
\section{Introduction}

People interact with objects 
routinely. 
Reconstructing 
human-object interaction (\HOI) 
in 3D is 
key for 
many applications, from 
robots to mixed reality. 
However, 
doing so 
from single images is challenging due to depth ambiguity, occlusions, and the diverse shape and appearance of objects.

There are methods that estimate 3D human bodies and methods that estimate 3D objects, but few that put these together.
Knowing the contacts %
between these %
could significantly improve joint reconstruction.
Our goal is to infer contact points on both humans and objects from single \inthewild images, and then use these contacts for jointly reconstructing humans and objects.
However, there is a lack of \inthewild training images paired with ground-truth contact labels for both 3D humans and objects.
Acquiring such data is challenging and existing methods do not scale.

The problem gets more challenging due to 
the complexity of real-world interactions.
Humans often contact multiple objects simultaneously; \eg, using a laptop while sitting on a club chair. 
Yet, current approaches treat contact prediction as simple binary classification; \ie detecting whether a body part is in contact with ``any'' object.
This simplified assumption fails to capture the rich semantic relationships of multi-object interactions.
To address this, %
we introduce a novel ``Semantic Human Contact" estimation task.
This %
involves predicting the contact points on 
the body 
related to 
a particular object 
given a single \inthewild image.

To tackle this, as well as 
to overcome data scarcity, %
we propose a new paradigm for scaling and reasoning in the wild.
Specifically, we observe that large Vision-Language Models (\VLMs)
can ``reason'' about \inthewild images because they are trained on internet-scale data and 
possess a broad visual knowledge about humans and their interactions with the world. 
We also observe that this knowledge can be re-purposed for novel tasks by fine-tuning these large models on small datasets. 
Thus, we exploit \VLMs for developing a novel framework, called \nameMethod. 

At the heart of \nameMethod lies a reasoning module based on a \VLM; see~\cref{fig:bridging_module_RSL}.
We enrich this %
with skills for 3D human-object ``understanding,'' by extending the \VLM with a \LORA~\cite{hu2021lora} adaptation layer. 
As a result, given a color image, this module can be ``asked'' to produce ``reasoning tokens'' that facilitate 3D contact localization. 

However, exploiting these tokens to localize contact is non-trivial.
A natural choice would be to employ a foundational ``localization'' model~\cite{sam} that 
takes these tokens as ``guidance,'' and highlights the 3D contacts. 
But there exists a key practical problem. 
Existing foundation %
models
inherently operate only in 2D space, while we need them to do so in 3D. 
To tackle this, we need to \mbox{re-cast} our problem so it is appropriate for %
2D foundation models.

To this end, we develop a novel ``\nameRepresentationLONG'' (\nameRepresentation) framework that has three main steps; see \cref{fig:bridging_module_RSL}: 
\highlight{(1)}~We render the 3D shape of a canonical 
\smplH~\cite{romero2022mano_smplh} 
body and an object as 2D images from multiple viewpoints. %
Regarding the object, a 3D mesh
is efficiently retrieved from a large-scale 3D database~\cite{objaverse} through \OpenShape~\cite{liu2023openshape}. 
\highlight{(2)}~We pass the above images into a foundation model to predict
2D contact masks for both the body and the object. 
\highlight{(3)}~We lift the predicted 2D contact points to 3D points via back-projection, \ie, performing the inverse of the first step. 

However, even after recasting 
the problem to 2D by rendering multi-view images, 
existing foundation models 
are still not 3D aware, \ie they treat each view independently, ignoring multi-view consistency.
This means contact detections in one view do not necessarily agree with ones in adjacent views.
To tackle this, just appending camera parameters to multi-view renderings is insufficient.
Instead, we build a novel ``Multi-view Localization'' model 
that we call 
\MVLoc.
This has two steps: 
\highlight{(1)} It transforms 
the ``reasoning token'' provided by the \VLM with the camera parameters 
used to render the multi-view images. 
\highlight{(2)} It ensures 
multi-view consistency, by lifting the inferred 2D contacts in each view to 3D and computing a 3D loss. 

\begin{figure}
    \centering
    \includegraphics[width=1.0 \columnwidth]{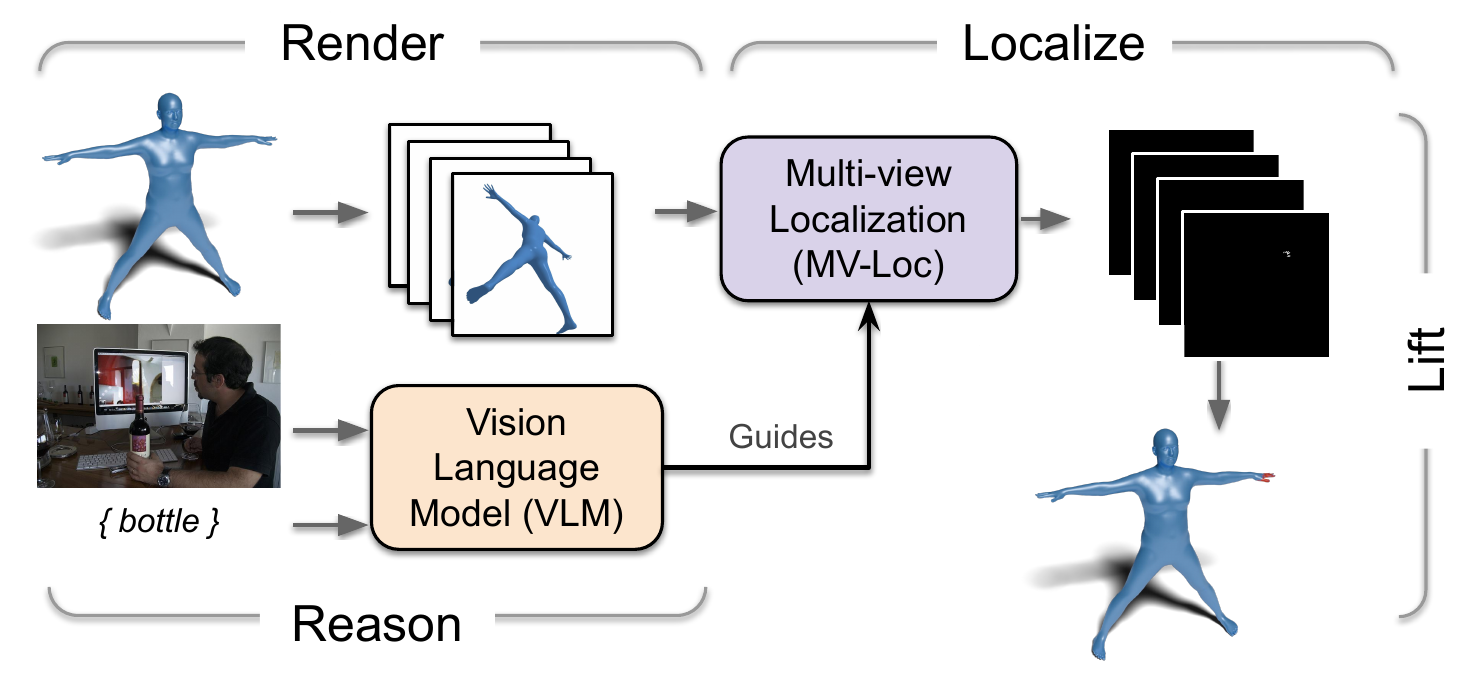}
    \vspace{-2.0 em}
    \caption{
        \qheading{Overview of \nameMethod} 
        Given a color image, our \orange{VLM} performs the core reasoning, and guides a novel \lilac{\mbox{MV-Loc}} model to localize contacts on both bodies and objects in 3D. 
        Here we show only the body; 
        for details, and object contact, %
        see \cref{fig:method}. 
    }
    \label{fig:bridging_module_RSL}
\end{figure}

Our method, \nameMethod, utilizes a \VLM in tandem with our novel multi-view localization model, \MVLoc, to perform 3D contact prediction for humans and objects.
See \cref{fig:teaser} for some examples. %
We quantitatively evaluate the efficacy of our method for \inthewild 3D 
contact prediction on bodies and on objects, using the \DAMON~\cite{tripathi2023deco} and \PIAD~\cite{yang2023piad} datasets, respectively. 
For bodies, we evaluate both for the traditional ``binary contact'' estimation, and for our new task of ``semantic contact'' estimation. 
For all tasks, we find that \nameMethod outperforms the prior work.

Finally, we demonstrate how \nameMethod's estimated contact improves 3D \HOI recovery from \inthewild images; this is a highly ill-posed task due to depth ambiguities and occlusions. 
To address this, we develop an optimization-based method that fits a \smplX body mesh and an \OpenShape-retrieved object mesh to the image, using \nameMethod's inferred contacts as constraints to anchor human and object meshes \wrt each other. 
To the best of our knowledge, this is the first %
approach for estimating  3D \HOI for \emph{\inthewild} images using \emph{inferred} contacts.

In summary, we make four key contributions:
\begin{enumerate} %
    \item 
    We build \nameMethod, a novel method that facilitates \HOI reconstruction from an \inthewild image by detecting 3D contacts on both bodies and objects. 
    \item 
    We demonstrate a way to minimize 
    reliance on 3D contact annotations via exploiting the broad visual knowledge of 
    Vision-Language Models. %
    \item 
    We build a novel ``Multi-view Localization'' model that helps in estimating contacts in 3D by transforming the reasoning of foundation models from 2D to 3D.
    \item 
    We introduce the novel ``Semantic Human Contact'' task for inferring body contacts conditioned on object labels.
\end{enumerate}
Our code and trained models
are available for research at 
\url{https://interactvlm.is.tue.mpg.de}.

\section{Related Work}

\subsection{Large Vision-Language Models}

Recent advancements in large language models (LLMs) have led to the development of multimodal models that integrate vision and language reasoning. 
Models like Flamingo~\cite{flamingo} and BLIP-2~\cite{blip2} use cross-attention mechanisms and visual encoders to align image features with text, supporting a variety of vision-language tasks. 
More recent works, such as \mbox{VisionLLM}~\cite{WangC23} and \mbox{Kosmos-2}~\cite{PengW23}, use grounded image-text data to enhance spatial understanding, while \mbox{GPT4RoI}~\cite{ZhangS23} introduces spatial box inputs for finer alignment. 
However, these models typically lack end-to-end segmentation capabilities.
To address this limitation, \mbox{LISA}~\cite{lai2024lisa} combines vision foundation segmentation models like \SAM~\cite{sam} with multimodal embeddings, enabling language-guided segmentation. 
\mbox{PARIS3D}~\cite{DoeS24} extends this approach to referential 3D segmentation by processing multi-view object renders through both \SAM and \mbox{LLaVA}~\cite{liu2023llava} for spatially-aware segmentation.

Taking inspiration from these approaches, we exploit 
a 
language-guided segmentation model for the task of predicting human and object contact in 3D. 
However, unlike PARIS3D, we process a single RGB image with \mbox{LLaVA} and multi-view renders of 
the human mesh and the object mesh 
with \SAM. 
Moreover, we introduce a feature-lifting technique that extends \mbox{LLaVA}'s 2D features into 3D using camera parameters, thus guiding \SAM's multi-view segmentations. 
This approach ensures multi-view consistency and efficiently predicts contact affordances, extending the use of multimodal models in 3D reasoning tasks focused on human-object interaction.

\subsection{3D Human and Object from Single Images}

Estimating 3D human pose and shape from a single image has evolved from optimization-based methods to learning-based approaches. 
Optimization-based methods fit parametric body models like \smpl~\cite{SMPL:2015}, \smplX~\cite{SMPL-X:2019}, or \mbox{GHUM}~\cite{ghum} to 2D cues such as keypoints~\cite{smplify}, silhouettes~\cite{nbf}, or segmentation masks~\cite{nbf}. 
Learning-based methods either regress body parameters from images or videos~\cite{Kocabas2021pare, dsr, VIBE:CVPR:2020, poco, joo2021eft, cliff, multi-hmr2024} or estimate non-parametric bodies as vertices~\cite{li2021hybrik, meshgraphormer}, implicit surfaces~\cite{mihajlovic2022coap, pifuhd}, or dense points~\cite{nlf}. 
Transformer-based methods~\cite{hmr2, osx, tokenhmr, sun2024aios} have further improved robustness. 

For 3D object reconstruction from a single image, regression-based methods predict geometry using meshes, voxels, or point clouds. 
Diffusion-based models~\cite{lrm} utilize large 3D datasets like \mbox{Objaverse}~\cite{objaverse} or 2D diffusion models~\cite{zero1to3, zero12345, magic3d, magic123} to guide reconstruction and optimization methods~\cite{antic2025sdfit} use render and compare.
Retrieval-based methods, such as \OpenShape~\cite{liu2023openshape} and \mbox{Uni3D}~\cite{zhou2023uni3d}, have demonstrated some robustness in cases with occlusions.

\begin{figure*}
    \centering
    \vspace{-0.5 em}
    \includegraphics[width=1.0\textwidth]{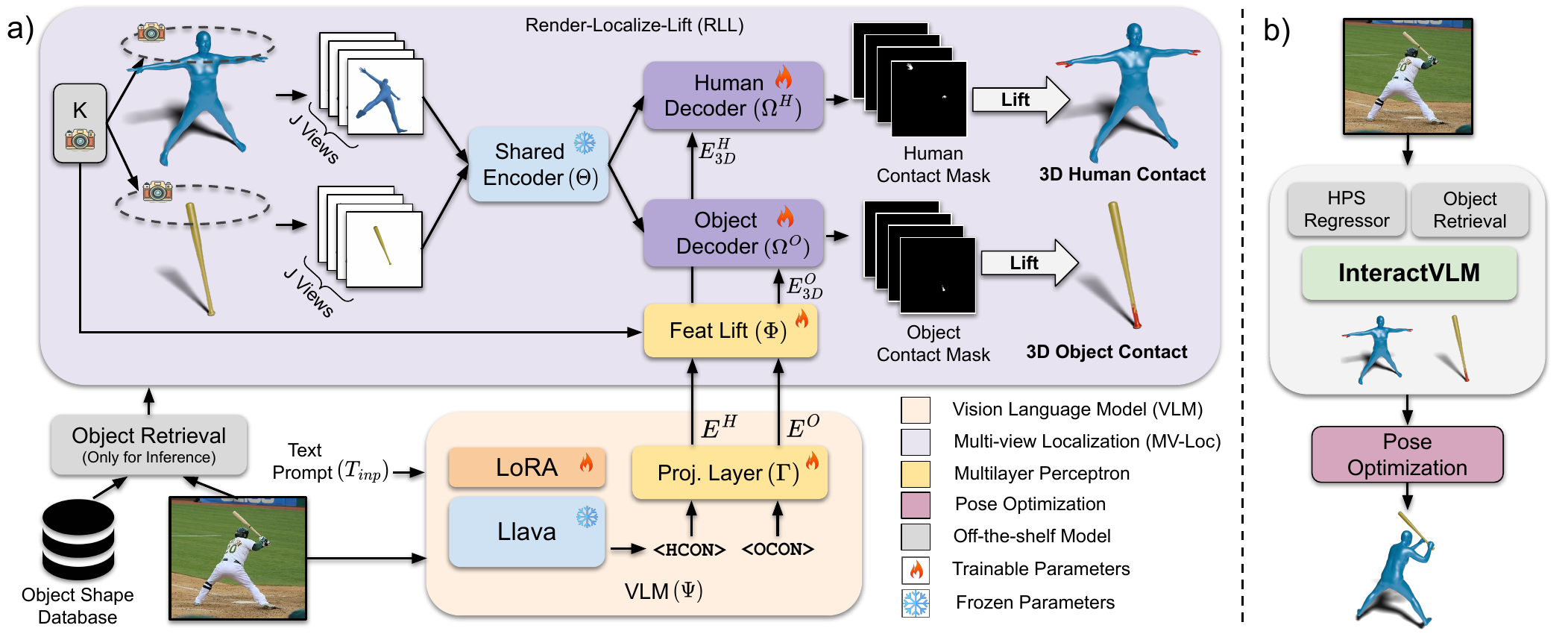}
    \vspace{-1.5 em}
    \caption{
                \qheading{Method overview}
                            Given a single \inthewild color image, our novel \nameMethod method estimates 3D contact points on both humans and objects (a). Then, we reconstruct a 3D human and object in interaction by exploiting these contacts (b). 
                            More specifically: 
                            \mbox{\qheading{(a)~Contact estimation}}
                            Given an image, $\image$, and 
                            prompt text, $\inputPrompt$, 
                            our \VLM, $\vlmNN$, produces contact tokens for humans and objects, \humanToken and \objectToken, which are projected ($\Gamma$) into feature embeddings, $\humanEmd$ and $\objectEmd$. 
                            These guide a ``Multi-View [contact] 
                            Localization'' %
                            model. 
                            This renders the 3D human and object geometry via cameras, $\camParams$, into multi-view 2D renders and passes these to encoder, $\imageEncoder$, while decoders, $\HmaskDecoder$, $\OmaskDecoder$, estimate and highlight 2D contacts in these renders. %
                            Then, 
                            the \FeatLift module, $\featLiftNN$, 
                            transforms 
                            the \VLM's features ($\humanEmd$, $\objectEmd$) 
                            to become 3D-aware 
                            ($\humanEmd_{3D}$, $\objectEmd_{3D}$) 
                            by exploiting the camera parameters, $\camParams$. 
                            A final module lifts the detected 2D contacts to 3D. 
                            \mbox{\qheading{(b)~3D \HOI reconstruction}}
                            For joint human-object reconstruction, we use \nameMethod's inferred contacts in an optimization framework. 
    }
    \label{fig:method}
\end{figure*}

\subsection{3D Human-Object Interaction}

Understanding 3D human-object interactions is essential for modeling realistic scenes. 
Early works focused on hand-object interactions, such as \mbox{ObMan}~\cite{hasson19_obman} and FPHA~\cite{FPA_CVPR2018}, with more recent studies like \mbox{ARCTIC}~\cite{fan2023arctic} and \mbox{HOLD}~\cite{fan2024hold} providing more detailed data and reconstruction for hands. 
For full-body interactions, initial studies involve interactions with scenes, as in \mbox{PROX}~\cite{PROX2019}, and 
with objects, 
as in \mbox{BEHAVE}~\cite{bhatnagar22behave}, \mbox{GRAB}~\cite{GRAB2020}, and \mbox{InterCap}~\cite{huang2024intercap}.
\mbox{BEHAVE} and \mbox{GRAB} use motion capture that are accurate but non-scalable, while \mbox{InterCap} uses multi-camera setups that are more scalable but less accurate.
Both of these approaches are limited in capturing diverse and realistic interactions.

As a proxy for 3D reconstruction, recent methods like \DECO~\cite{tripathi2023deco} and HOT~\cite{chen2023hot} infer contact on body meshes and image pixels, respectively, for which contact annotations are crowdsourced.
Predicting contact on objects is more challenging due to varying object shapes, 
while also currently there exists no dataset 
of 
object contact for in-the-wild images. %
Thus, we estimate 3D object affordances as a proxy for contact. 
\mbox{3D-AffordanceNet}~\cite{affordancenet3d} introduces affordances that are not grounded in images, capturing the likelihood of humans interacting with specific parts of an object for a given affordance (e.g., ``sit on chair''). \PIAD~\cite{yang2023piad} curates RGB images depicting object affordances and trains a network to estimate them. 
\LEMON~\cite{yang2023lemon} extends 
object affordance prediction of \PIAD to include human contact estimation.
However, these methods require human contact vertices 
\emph{paired} with
object affordances for training, limiting the number of categories they handle 
to 21 ones. 
In contrast, we learn from \emph{unpaired} human and object interaction data 
enabling human interaction reasoning for 80 categories and object affordance prediction for 32 categories. 

\subsection{Joint 3D Human-Object Reconstruction}

Joint reconstruction of humans and objects in 3D has been 
tackled with 
both regression and optimization methods. 
Regression-based methods directly predict 3D human-object meshes, as in \HDM~\cite{xie2023template_free} and \CONTHO~\cite{nam2024contho}, while other methods %
first predict contact points and then use test-time optimization to fit human and object meshes, as in \CHORE~\cite{xie2022chore} and \PHOSA~\cite{zhang2020phosa}. 
Since regression methods rely on limited training data, optimization methods are preferred for in-the-wild scenarios, such as \PHOSA~\cite{zhang2020phosa}. 
Optimization-based methods either assume known contacts or infer contacts to fit meshes to the image, but their success heavily relies on 
the quality of contact. 

Our method improves upon these by providing more accurate contact predictions, which in turn facilitate better fitting of human and object meshes to image evidence, improving the realism and accuracy of 3D human-object reconstructions from single images. 

\section{Method}

\subsection{Input Representation}
\label{sec:initialization}

Given an %
image, $\image \in \mathbb{R}^{H \times W \times 3}$, \nameMethod estimates 3D contacts for both human bodies and objects. 

The human is 
represented by a 
\smplh \cite{romero2022mano_smplh} %
3D body mesh, $\human$, with vertices 
$\vertices^\human \in \mathbb{R}^{10475\times3}$.  
The body is posed in a canonical ``star'' shape (see details in~\cref{sec:method_MVLoc}).
The human contacts are binary per-vertex labels, $\hcontact \in \{0, 1\}$. 

The object is represented by a 3D point cloud (or mesh), $\object \in \mathbb{R}^{N\times3}$, with $N$ points. %
As there are no datasets of natural images paired with 3D contacts for objects, 
we use a large-scale 3D affordance dataset~\cite{yang2023piad}, instead, as affordances are closely related to contact. %
Specifically, they represent the likelihood of contact on 3D object areas for various purposes. 
Therefore, for objects we use the terms ``affordance'' and ``contact'' interchangeably.
The object contacts are %
continuous per-point values, $\ocontact \in [0, 1]$.
During inference, %
we %
retrieve a 3D object shape from a large database~\cite{objaverse} via \OpenShape~\cite{liu2023openshape} conditioned on image~$\image$.

\subsection{Overview of \nameMethod}
The biggest challenge for learning %
3D contact prediction %
in the wild 
is the limited 3D contact data 
for humans and objects. 
To go beyond existing limited datasets, 
we introduce a novel method, called \nameMethod, that harnesses the commonsense knowledge of large vision-language models. 
Specifically, \nameMethod (Fig.~\ref{fig:method}) has two main components: a Vision Language Model (\VLM) and a novel Multi-View contact Localization model (\MVLoc). 
\MVLoc highlights parts that are in contact for both humans and objects with the \VLM's guidance. 
The input to the \VLM (\cref{sec:method_VLM}) is an image, $\image$, and a prompt, $\inputPrompt$, that asks the \VLM to 
detect contact. 
The input to \MVLoc (\cref{sec:method_MVLoc}) is the 3D geometry of humans and objects, $\human$ and $\object$, respectively. %

\subsection{Interaction Reasoning through \VLM}
\label{sec:method_VLM}
The \VLM module, $\vlmNN$, conducts the core interaction reasoning. 
It takes an input image, $\image$, and prompt text, $\inputPrompt$, and outputs a prompt text, 
$\outputPrompt = \vlmNN(\image, \inputPrompt)$. 
Inspired by the recent 
\LISA~\cite{lai2024lisa} model, we expand the \VLM's vocabulary with two specialized %
tokens, \humanToken and \objectToken, 
for 
contact information for humans and objects, respectively. 

To denote contact, $\vlmNN$ produces a prompt that includes the above tokens. 
To aid \MVLoc in localizing contact, we extract the last-layer embeddings of the \VLM corresponding to these tokens and pass them through a projection layer, $\projLayerNN$, to obtain the feature embeddings, $\humanEmd$ or $\objectEmd$. 
Let $ T_{\text{gt}} $ be the ground-truth text, and $ T_{\text{pred}} $ be the 
predicted one. 
Then, our token-prediction loss %
is defined as a cross-entropy loss:
\begin{equation}
    \tokenLoss = - \textstyle\sum_{i=1}^{N}   (  T_{\text{gt}}^{(i)} \cdot \log(T_{\text{pred}}^{(i)})  ) .
\end{equation}

\subsection{Interaction Localization through \MVLoc}
\label{sec:method_MVLoc}
We develop a novel \MVLoc module %
that has 
a shared image encoder, $\imageEncoder$, and separate decoders, $\HmaskDecoder$ and $\OmaskDecoder$, for humans and objects.
\MVLoc 
performs contact localization by %
using
a novel ``\nameRepresentationLONG'' (\nameRepresentation) 
framework.
To this end, it 
takes three steps; 
(1)~rendering the 3D shape of humans and objects in 2D, 
(2)~predicting 2D contact maps for both of these, %
and 
(3)~lifting the 2D contact maps to 3D. 

\zheading{\nameRepresentation step~\#1:~\Render 3D\textrightarrow 2D} 
The input %
is human and object geometry, namely $\humanMesh$ and $\objectMesh$, respectively. 
Both serve as a ``canvas'' for ``painting'' detected contacts on these. 
The body has a default \smplH shape  in a canonical star-shape pose to minimize self-occlusions 
when rendering. %
The object geometry (initialized in \cref{sec:initialization}) 
is normalized to a unit sphere.
Each geometry is rendered from %
$\jviews$ fixed views with camera parameters, $\camParams$,  to form multi-view renderings, 
$\inputRenderOH = \{\inputRender_j\}_{j=1}^\jviews$, 
so that 
the entire 3D geometry is captured.
Since our geometries do not have texture, we color 
the meshes with normals and point clouds using the NOCS map~\cite{nocs}.
This enhances cross-view correspondence, 
making renderings resemble real images 
to 
image encoder, 
$\imageEncoder$.

\zheading{\nameRepresentation step~\#2:~Localize in 2D}
The rendered geometry, $\inputRenderOH$, is first sent to the image encoder, $\imageEncoder$, and then passed to %
decoders, so that %
the final contact masks, $\humanMask$ and $\objectMask$, get highlighted on it. 
However, \MVLoc requires spatial and contextual cues for highlighting the contact region. 
To this end, we use the feature embeddings (\cref{sec:method_VLM}), \ie $\humanEmd$ and $\objectEmd$ to guide the contact localization. %

However, since the \VLM reasons in 2D, these features are not 3D aware. 
This is a problem, because \MVLoc needs 3D awareness to localize contact consistently across multi-view renderings. 
Thus, we transform the features to ``lift'' them to ``3D'' 
to 
better guide 
multi-view localization.

In detail, we design a lifting network, $\featLiftNN$, 
which takes 
camera parameters, $\camParams$, and 
the 2D $\humanObjectEmd$, and lifts the latter
to 3D as $\humanObjectEmdLift = \featLiftNN(\humanObjectEmd, \camParams)$.
Contact masks %
are defined as:
\vspace{-0.35 em}
{%
\begin{equation}
    \humanObjectMask = \HOmaskDecoder \bigl( \imageEncoder(\inputRenderOH), \humanObjectEmdLift \bigr)
    \text{.}
\end{equation}
}

\noindent
Below, the 
$\human$ and $\object$ superscripts are dropped for brevity. %
We calculate losses only on ``valid'' regions, \ie, the areas within the outline of rendered geometry; we denote this as $\validmask$. 
To encourage overlap between 
the predicted masks, $\validmask$, and the ground-truth ones, $\validmaskGT$,
particularly in sparse contact regions,
we use a focal-weighted BCE loss and a Dice loss:
\vspace{-0.35 em}
{%
\begin{equation}
    \bceLoss = -\alpha (1 - \validmaskP)^{\gamma} \log(\validmaskP) - (1 - \alpha) \validmaskP^{\gamma} \log(1 - \validmaskP) \text{,}
\end{equation}
\begin{equation}
    \diceLoss = 1 - \frac{2 \sum \validmask \cdot \validmaskGT + \epsilon}{\sum \mask + \sum \validmaskGT + \epsilon} \text{,}
\end{equation}
}

\noindent
where $\maskP$ is the predicted mask probability, $\alpha$ controls class balance, $\gamma$ adjusts the focus on hard examples,
and \(\epsilon\) is a residual %
term to prevent division by zero.

\zheading{\nameRepresentation step~\#3:~\Lift 2D\textrightarrow 3D}
To lift the inferred 2D contact points to 3D points, we follow the inverse of step~\#1.
Normally, 2D points backproject to 3D lines due to depth ambiguities. 
In our case the lines intersect with the known 3D geometry that produced the multi-view renders. 
So, 2D points are lifted to %
3D points, 
and by extension 
2D contact masks, $\humanObjectMask$, are lifted to %
3D contact areas, %
$\ohcontact$.

We use a human contact loss, $\hCLoss$, that combines a focal loss with sparsity regularization,
to encourage precise true-positive predictions 
in ``valid'' contact regions while discouraging false positives in non-contact areas:
\begin{equation}
    \hCLoss = \alpha (1 - \hcontactP)^{\gamma} \log(\hcontactP) + \lambda \|\hcontact\|_1,
\end{equation}
where $\hcontactP$ is the contact probability, and 
$\lambda$, %
$\alpha$ and $\gamma$ are 
scalar weights. %
We also use an object contact loss, $\oCLoss$, that 
combines a 
Dice and Mean Squared Error (MSE) loss:
\begin{equation}
    \oCLoss = \diceLoss(\ocontact, \ocontactGT) + \beta \|\ocontact - \ocontactGT \|_2^2,
\end{equation}
where \(\beta\) is a weighting factor, and $\ocontactGT$ denotes the ground-truth 3D contacts for objects.

\subsection{Implementation Details}
\label{sec:implementation}

\zheading{Architecture}
We use \LLAVA~\cite{liu2023llava} as our \VLM and \SAM \cite{sam} for our \MVLoc, 
with weights pre-trained by \LISA~\cite{lai2024lisa} for segmentation~\cite{lai2024lisa}. 
The feature-lifting network, $\featLiftNN$, contains a spatial-understanding network (two fully-connected layers of size 128 with \mbox{ReLU} activation) followed by view-specific 
(256-dimensional) transformations and a sigmoid activation. 
For 3D contact prediction, 
our \MVLoc model 
converts 2D masks %
to 3D contact points via  
2D-to-3D pixel-to-vertex mappings that are precomputed during \MVLoc's rendering step. 
For details, see \supmat

\zheading{Training}
To efficiently fine-tune our \VLM, we employ \LORA~\cite{hu2021lora} with rank 8.
The separate decoders for human and object contact prediction are trained without \LORA, %
while keeping the image encoder frozen.
For training, we use DeepSpeed~\cite{deepspeed} with mixed precision training (bfloat16), and batch size of 8. %
We train on 4 Nvidia-A100 GPUs for 30 epochs. 
For more details, please refer to \supmat

\zheading{Datasets}
We focus %
on 
two tasks, \ie, 3D human contact and 3D object affordance prediction, 
using two \inthewild datasets, \ie, \DAMON~\cite{tripathi2023deco} and \PIAD~\cite{yang2023piad}, respectively. 
For human contact we train and evaluate on \DAMON~\cite{tripathi2023deco}. %
For 3D object affordances, we train and evaluate on \PIAD~\cite{yang2023piad}. %
We find that exploiting textual descriptions for the contacting body parts, and the object type in contact, helps training. 
Similarly, adding Visual Question-Answering (\VQA) data generated by \GPT for the training images also helps. %
For details, see \supmat

Unlike LEMON~\cite{yang2023lemon}, which requires paired human-object geometry for training using the \mbox{3DIR} dataset~\cite{yang2023lemon}, we use unpaired data.
This enables us to scale for many human and object categories not addressed by prior work. %
For the final joint human-object reconstruction task, we combine \DAMON~\cite{tripathi2023deco}, \PIAD~\cite{yang2023piad}, \mbox{3DIR}~\cite{yang2023lemon} and all textual descriptions about body parts, contact type and \HOI.

\zheading{Evaluation metrics}
For human contact prediction, following~Tripathi~\etal~\cite{tripathi2023deco}, we report the F1, precision, and recall scores using a threshold of 0.5, and a geodesic distance measuring spatial accuracy. %
For object contact prediction, following Yang~\etal~\cite{yang2023piad}, we report the Similarity (SIM), Mean Absolute Error (MAE), Area Under ROC Curve (AUC), and average Intersection over Union (IOU). %

\section{Experiments}

\begin{table}
    \centering
    \vspace{-0.5 em}
    \scriptsize
    \resizebox{0.95 \columnwidth}{!}{
        \begin{tabular}{l|cccc}
            \toprule
            Method & F1 & Precision  & Recall  & Geodesic \\
             & (\%) $\uparrow$ & (\%) $\uparrow$ & (\%) $\uparrow$ & (cm) $\downarrow$ \\
            \midrule
            $\POSA^\text{{\scriptsize{\PIXIE}}}$~\cite{POSA_Hassan_2021_CVPR}~\cite{pixie}  & 31.0 & 42.0 & 34.0 & 33.00 \\
            \BSTRO~\cite{RICH}             & 46.0 & 51.0 & 53.0 & 38.06 \\
            \DECO~\cite{tripathi2023deco}  & 55.0 & 65.0 & 57.0 & 21.32 \\
            \midrule
            \textbf{\nameMethod} & \textbf{75.6} & \textbf{75.2} & \textbf{76.0} & \textbf{2.89} \\
            \bottomrule
    
        \end{tabular}
    }
    \caption{
                Evaluation for ``Binary Human Contact'' prediction on the \DAMON dataset~\cite{tripathi2023deco}. 
                We compare our \nameMethod model (trained only \CR{for} this task) with the state of the art. 
    } 
    \label{tab:binary_human_contact}
\end{table}

\begin{figure*}
    \centering
    \includegraphics[width=0.99\textwidth]{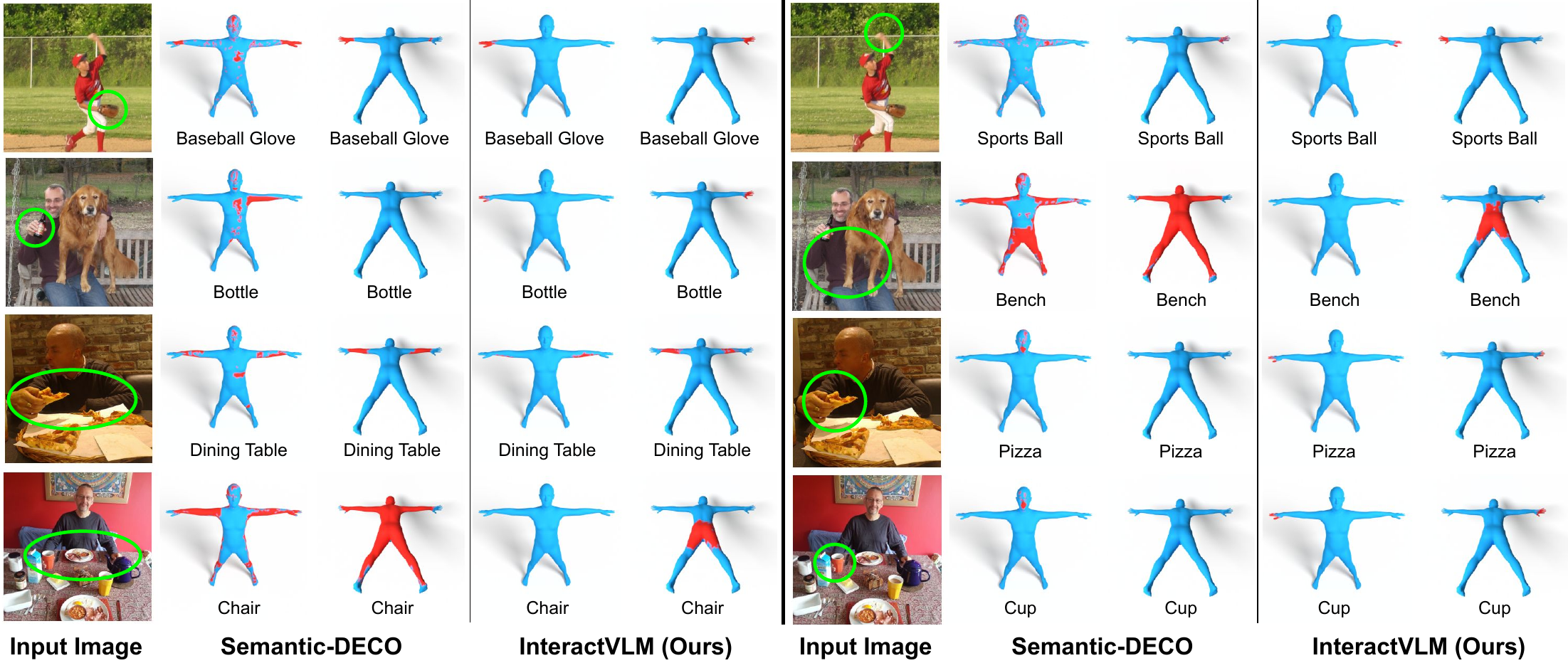}
    \vspace{-0.5 em}
    \caption{
                \qheading{``Semantic Human Contact'' estimation~(\cref{sec:semantic_human_contact})} 
                Given an image and an object label, \nameMethod infers body contacts for this object. 
                \nameMethod outperforms a Semantic-\DECO~\cite{tripathi2023deco} baseline. 
                Objects are shown in green circles, and contacts as red patches.
    }
    \label{fig:qualitative-body-contacts}
\end{figure*}

\subsection{``Binary Human Contact'' Estimation}
\label{sec:experiments-binary-contact}
This task refers to 
estimating contact areas on the body via binary classification of its vertices, 
ignoring 
the number or type of objects involved. 
We train and evaluate on the \DAMON~\cite{tripathi2023deco} dataset, and report results in~\cref{tab:binary_human_contact}. 

\nameMethod 
significantly outperforms all previous methods achieving a 20.6\% improvement in F1 score.
Although, here, \nameMethod is trained on the same data as \DECO, it goes beyond this 
by harnessing the commonsense knowledge of a large foundation model.

In \supmat~we also evaluate on the \mbox{3DIR}~\cite{yang2023lemon} dataset and compare with the \LEMON method~\cite{yang2023lemon}. 
Even though \LEMON uses paired human-object data, \nameMethod performs %
on-par with it despite training on human-only data.
We also evaluate \nameMethod's performance for binary human contact prediction across different body parts; for more details please refer to \supmat

\begin{figure}
    \vspace{-0.5 em}
    \centering
    \includegraphics[width=0.95 \columnwidth]{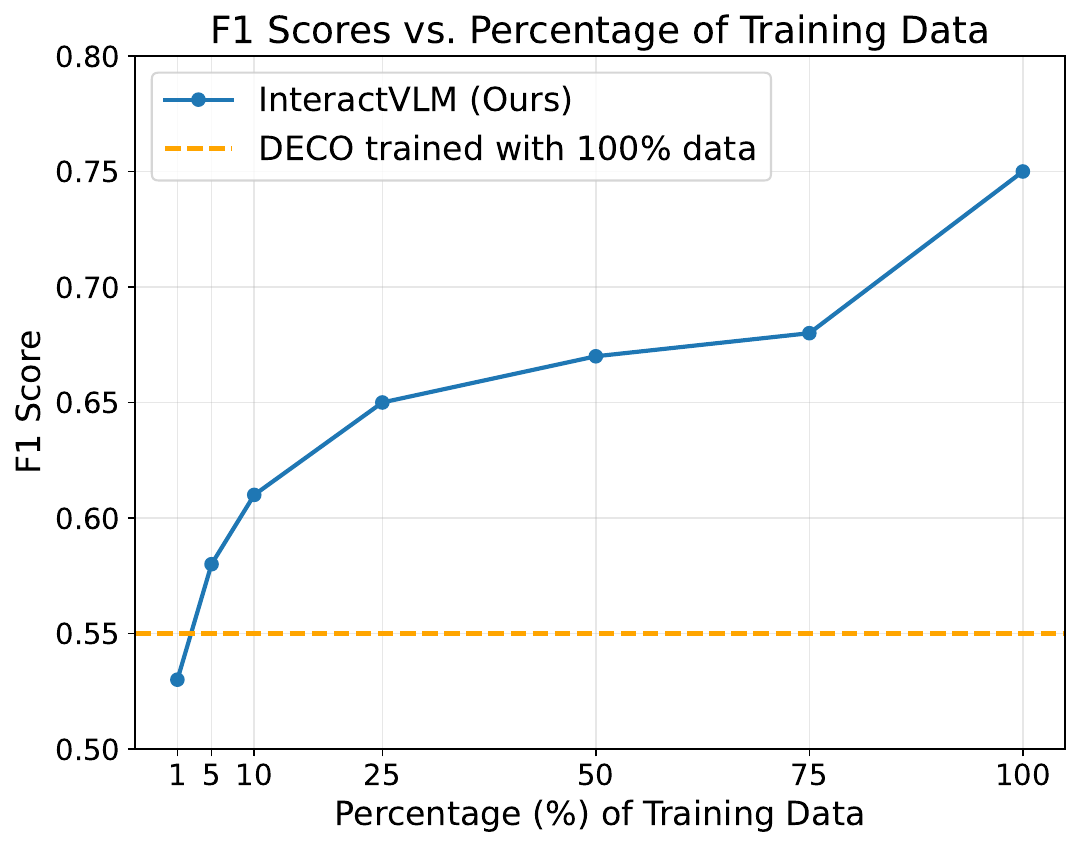}
    \caption{
        \qheading{\nameMethod's reliance on 3D annotations} 
        We evaluate performance for ``binary human contact'' (F1 score, Y-axis) for models trained on a varying percentage of \DAMON~\cite{tripathi2023deco} training data (X-axis). 
        The \DECO baseline trains on 100\% of \DAMON. 
        Instead, \nameMethod trains on a varying (smaller) portion of this dataset. %
        Yet, it achieves a significantly higher performance, 
        by leveraging the broad visual knowledge of foundation models.
    }
    \label{fig:f1_vs_training_data}
\end{figure}

\begin{table}
    \centering
    \resizebox{\columnwidth}{!}{
        \begin{tabular}{l|cccc|cccc}
            \toprule
            & \multicolumn{4}{c|}{\semanticDECO~\cite{tripathi2023deco} (Baseline)} & \multicolumn{4}{c}{\textbf{\nameMethod}} \\
            \cmidrule{2-5} \cmidrule{6-9}
            Object      & F1 & Prec. & Rec. & Geo & F1 & Prec. & Rec. & Geo \\
            Categories  & (\%) $\uparrow$ & (\%) $\uparrow$ & (\%) $\uparrow$ & (cm) $\downarrow$ & 
            (\%) $\uparrow$ & (\%) $\uparrow$ & (\%) $\uparrow$ & (cm) $\downarrow$ \\
            \midrule
            \rotatebox[origin=c]{0}{Accessory} & 40.1 & 30.7 & \textbf{75.6} & 21.88 & \textbf{61.1} & \textbf{72.9} & 65.6 & \textbf{6.91} \\
            \rotatebox[origin=c]{0}{Daily Obj}  & 26.5 & 20.1 & 52.3 & 60.34 & \textbf{68.6} & \textbf{71.4} & \textbf{78.0} & \textbf{7.46} \\
            \rotatebox[origin=c]{0}{Food}      & 11.7 & 19.4 & 12.9 & 49.61 & \textbf{66.4} & \textbf{66.1} & \textbf{77.9} & \textbf{7.17} \\
            \rotatebox[origin=c]{0}{Furniture} & 24.5 & 15.8 & \textbf{83.7} & 29.17 & \textbf{60.5} & \textbf{64.2} & 60.1 & \textbf{6.21} \\
            \rotatebox[origin=c]{0}{Kitchen}   & 27.7 & 24.7 & 37.2 & 52.34 & \textbf{71.8} & \textbf{71.3} & \textbf{81.1} & \textbf{7.61} \\
            \rotatebox[origin=c]{0}{Sports}    & 36.4 & 30.4 & 80.1 & 79.21 & \textbf{77.9} & \textbf{76.7} & \textbf{83.2} & \textbf{7.98} \\
            \rotatebox[origin=c]{0}{Transport} & 52.0 & 39.1 & 93.7 & 31.78 & \textbf{77.8} & \textbf{80.5} & \textbf{78.9} & \textbf{7.97} \\
            \bottomrule 
        \end{tabular}
    }
    \caption{
                Evaluation for ``Semantic Human Contact'' prediction on the \DAMON~\cite{tripathi2023deco} dataset. 
                For results on each class, see \supmat 
                The ``Semantic-\DECO'' baseline extends \DECO for our new task. 
    }
    \label{tab:results-semantic-contact-Human}
\end{table}

\begin{table}
    \vspace{-0.5 em}
    \centering
    \resizebox{\columnwidth}{!}{
        \begin{tabular}{l|cccc|cccc}
            \toprule
            & \multicolumn{4}{c|}{PIAD-Seen~\cite{yang2023piad}} & \multicolumn{4}{c}{PIAD-Unseen~\cite{yang2023piad}} \\
            \cmidrule{2-5} \cmidrule{6-9}
            Methods & SIM & AUC  & aIOU  & MAE  & SIM  & AUC  & aIOU & MAE  \\
                    & (\%) $\uparrow$ & (\%) $\uparrow$  & (\%) $\uparrow$  &  $\downarrow$  & (\%) $\uparrow$  & (\%) $\uparrow$  & (\%) $\uparrow$ & $\downarrow$  \\
            \midrule
            PMF~\cite{PMF}             & 42.5 & 75.05 & 10.13 & 1.41 & 33.0 & 60.25 & 4.67 & 2.11 \\
            ILN~\cite{ILN}             & 42.7 & 75.84 & 11.52 & 1.37 & 32.5 & 59.69 & 4.71 & 2.07 \\
            PFusion~\cite{PFusion}     & 43.2 & 77.50 & 12.31 & 1.35 & 33.0 & 61.87 & 5.33 & 1.93 \\
            XMF~\cite{XMF}             & 44.1 & 78.24 & 12.94 & 1.27 & 34.2 & 62.58 & 5.68 & 1.88 \\
            IAGNet~\cite{yang2023piad} & 54.5 & 84.85 & 20.51 & 0.98 & 35.2 & 71.84 & 7.95 & 1.27 \\
            \midrule
            \textbf{\nameMethod}  
                                        
                                        & \textbf{62.7} & \textbf{86.47} & \textbf{21.20} & \textbf{0.81} & \textbf{41.4} & \textbf{75.45} & \textbf{8.50} & \textbf{0.99} \\
            \bottomrule
        \end{tabular}
    }
    \caption{
                Evaluation for 
                ``Object Affordance Prediction'' 
                on the \PIAD \cite{yang2023piad} dataset. 
                We compare our \nameMethod model (trained only \CR{for} this task) with the state of the art. 
    }
    \label{tab:object_affordances}
\end{table}

\subsection{``Semantic Human Contact'' Estimation}
\label{sec:semantic_human_contact}
In real-life interactions, 
multiple objects can be contacted by different body areas concurrently. 
Thus, we introduce a novel task called ``Semantic Human Contact'' prediction. 
We evaluate on \DAMON~\cite{tripathi2023deco} and report results in~\cref{tab:results-semantic-contact-Human};
for a finer-grained version of this table, see \supmat  %

To establish a baseline, we adapt the \DECO model~\cite{tripathi2023deco} that detects binary contacts %
and turn it into a multi-class %
prediction model, called \semanticDECO.
Due to \DAMON's limited training data, this has poor performance. %
Instead, as discussed in \cref{sec:experiments-binary-contact}, \nameMethod learns effectively from this data, by 
also leveraging the commonsense of foundation models. %
Thus, it significantly outperforms \semanticDECO. %
Qualitative results reflect this finding; see~\cref{fig:qualitative-body-contacts}.
Our model captures detailed, accurate contact regions, whereas \semanticDECO often highlights false-positive areas that 
differ from the actual  contact regions.

\subsection{Object Affordance Prediction}
\label{sec:experiments-object-affordance}
We evaluate the performance of our \nameMethod model on predicting object affordances. 
This involves identifying regions on objects %
of possible %
contact for a certain interaction intent, such as ``sitting'' on a chair or ``moving'' it. 
We train and evaluate on the \PIAD~\cite{yang2023piad} dataset. 
We compare against \SOTA methods, and report results in~\cref{tab:object_affordances}. 

Note that we evaluate 
on object instances that are both seen during training (``\PIAD-Seen'' column) and unseen (``\PIAD-Unseen'' column). 
Our model demonstrates a significant improvement over \SOTA methods in terms of similarity (SIM), area under the curve (AUC), and mean absolute error (MAE) for both seen and unseen objects.
Thus, 
our method is effective not only for estimating human contact (\cref{sec:experiments-binary-contact,sec:semantic_human_contact}), but also for object affordances. 

\subsection{Reliance on 3D Annotations}
To analyze \nameMethod's efficiency regarding %
3D supervision, we 
train multiple versions of it 
with varying amounts of training data from the \DAMON dataset, 
and report performance in \cref{fig:f1_vs_training_data}. 
Remarkably, \nameMethod achieves an F1 score of $0.53$ with just 1\% of the data, nearly matching the $0.55$ F1 score of \DECO that uses 100\% of the data. %
With only 5\% of the data, \nameMethod surpasses \DECO, reaching an F1 score of $0.58$.
This performance gap widens as training data increases, ultimately achieving $0.75$ F1 score with 100\% of the data.
This is a compelling finding, 
highlighting \nameMethod's efficient use of 3D supervision by leveraging the rich visual understanding of foundation models.
Note that the strong performance with limited training data 
has high practical value, 
because 
obtaining 3D annotations is expensive.

\zheading{Ablations}
For a study on the influence of \nameMethod components, including mask resolution, \MVLoc variants, loss functions, training data, the influence of the \VLM, and the effect of text prompts, please refer to \supmat

\begin{figure*}
    \vspace{-1.5 em}
    \centering
    \includegraphics[width=0.99 \textwidth]{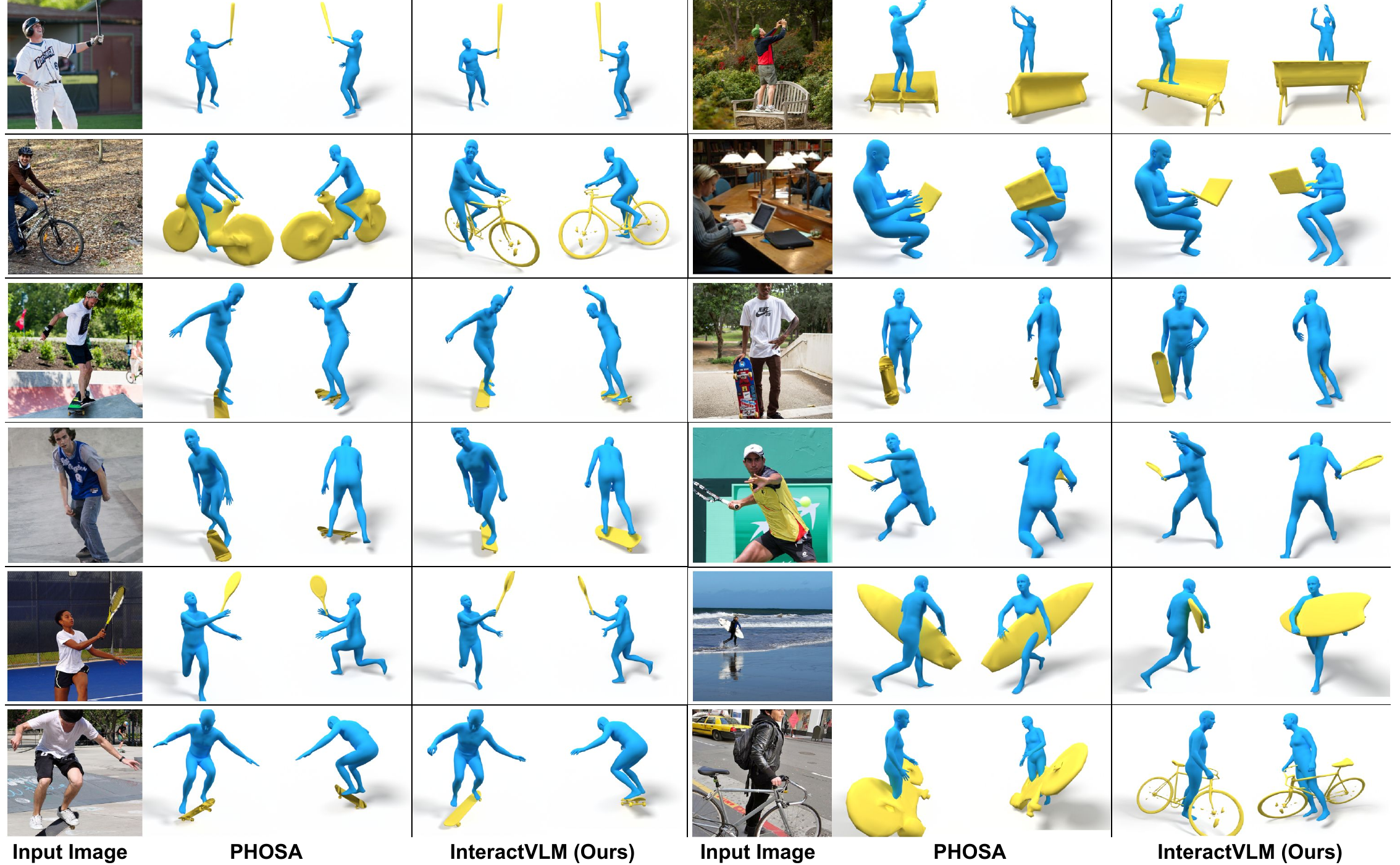}
    \vspace{-0.5
    em}
    \caption{
                \qheading{3D \HOI reconstruction (\cref{sec:downstream-qualitative})} 
                We build an optimization method %
                that fits a \smplX body and \OpenShape-retrieved object to an \inthewild image. 
                We evaluate against the \SOTA method \PHOSA~\cite{zhang2020phosa}. 
                Reconstruction is guided by \nameMethod-inferred contacts.
    }
    \label{fig:qualitative-3D-HOI}
\end{figure*}

\section{Joint Human-Object Reconstruction}
\label{sec:downstream-method}
We demonstrate the usefulness of 3D contacts inferred by \nameMethod for reconstructing a 3D human and object in interaction from a single \inthewild image, $I$. 

\zheading{Initializing 3D body pose \& shape, object shape}
We use %
\OSX~\cite{osx} %
to estimate a 3D 
\smplX body mesh, $\human$, 
and \OpenShape~\cite{liu2023openshape} to  
retrieve a 3D object mesh, $\object$, from the 
Objaverse~\cite{objaverse}
database that best matches the image.

\zheading{Initializing 3D object pose} 
We apply \nameMethod on image $\image$ to predict 3D body and object contact vertices, $\hcontact, \ocontact$. %
Then, we solve for object pose, $\{ \objectRot, \objectTransl \}$ by applying the ICP~\cite{icp} algorithm to snap the 3D object contact points, $\ocontact$, onto body ones, $\hcontact$. 
To avoid false correspondences, %
the 3D normals of contact points %
must be compatible, \ie, %
should have similar angles but opposite directions.

\zheading{Optimizing 3D object pose} 
Given the above initialization, %
we optimize over object rotation, $\objectRot$, translation, $\objectTransl$, and scale, $\objectScale$, via render-and-compare by minimizing: %
{%
\begin{align}\label{eq:energy}
    E &= E_{M} + \lambda_{C} E_{C}
    \text{,} \\
    E_M &= IoU(\rendMask, \gtMask) + ||\rendMeanMaskPxl - \gtMeanMaskPxl||_2                    \text{,}
    \\
    E_C &= \frac{1}{|\hcontact| |\ocontact|} 
            \sum_{i \in |\human|} 
            \sum_{j \in |\object|}    
            \hcontact_i \ocontact_j 
            \| \vertices^\human_i - \vertices^\object_j \|_2 \text{,}
\end{align}}where 
$E_M$ is a mask loss, 
$E_C$ is a contact loss, 
$IoU$ is intersection-over-union, 
$\rendMask, \gtMask$ are hypothesis and ground-truth masks, respectively, 
$\rendMeanMaskPxl, \gtMeanMaskPxl$ are the corresponding mean mask pixels, 
$|\hcontact|$, $|\ocontact|$ are the number of contact vertices on the human and object, 
$|\human|$, $|\object|$ are the number of mesh vertices, 
$\vertices^\human_i$, $\vertices^\object_j$ are the $i$-th and $j$-th human and object vertices, while %
$\ocontact_i$, $\hcontact_j$ indicate whether vertices are in contact or not.
Given the image, we extract the object mask, $\gtMask$, via \SAM~\cite{sam}, and a depth map, $D$, via Depth Pro~\cite{Bochkovskii2024:arxiv}. %

Intuitively, we use the $E_M$ loss to align the 3D object to the image, while 
in $E_C$ the predicted 3D contacts ``anchor'' the 3D object onto the body so they interact realistically. 
We perform iterative optimization via \mbox{Adam}~\cite{adam}. 
\OSX produces reasonable bodies, so we keep these fixed. %
The object is updated in each iteration -- we 
render depth, $\rendDepth$, and masks, $\rendMask$, via a \mbox{PyTorch3D}~\cite{ravi2020pytorch3d} differentiable renderer.

\zheading{Qualitative Results}
\label{sec:downstream-qualitative}
We reconstruct 3D human-object interaction from an image.
We show results in~\cref{fig:qualitative-3D-HOI}, and compare with \PHOSA~\cite{zhang2020phosa}, the most related \SOTA method. 
\nameMethod's reconstructions look more realistic. 
Note that \PHOSA uses handcrafted contacts on humans and bodies.
Instead, \nameMethod infers 3D contacts on both 
human bodies and objects
from the 
single \inthewild image.
These play a crucial role for guiding 3D reconstruction under occlusions and depth ambiguities.

\zheading{Perceptual Study}
There exists no \inthewild dataset with 3D ground truth for \HOI, 
so we conduct a perceptual study via Amazon Mechanical Turk for evaluation. 
Specifically, we evaluate the realism of our reconstructions %
against ones of \PHOSA. %
We randomly select 55 images for which \PHOSA has %
handcrafted contact annotations. %
For each image, participants are shown (with random swapping) reconstructions generated by our method and by \PHOSA, and are asked to select the one that best represents the image.
Our reconstructions 
are preferred $62\%$ of the time.

\section{Conclusion}

We develop \nameMethod, a novel method for 
estimating 3D contacts on both humans and objects from a single natural image. 
\nameMethod has a reduced reliance on expensive 3D contact annotations for training, by leveraging the broad knowledge of Vision-Language Models. 
Specifically, we introduce a novel ``\nameRepresentationLONG'' (\nameRepresentation) framework and a novel multi-view localization model (\MVLoc) to adapt 2D foundation models for 3D contact estimation.
We outperform existing work on contact estimation and introduce a new ``Semantic Human Contact'' estimation task for inferring body contacts conditioned on object labels.
This goes beyond traditional binary contact estimation, 
which fails to capture rich semantic relationships of multi-object interactions.
Last, we develop 
the first 
approach that uses inferred contact points on both bodies and objects for joint 3D reconstruction from single \inthewild images.

\pagebreak

{\small
\noindent
\textbf{Acknowledgments:}
We 
thank Alp\'{a}r Cseke for 
his assistance with evaluating joint human-object reconstruction.
We also thank Tsvetelina Alexiadis and Taylor Obersat for MTurk evaluation,
Yao Feng, Peter Kultis and Markos Diomataris
for their valuable feedback and
Benjamin Pellkofer for IT support.
SKD is supported by the International Max Planck Research School for Intelligent Systems (IMPRS-IS). 
The UvA part of the team is supported by an ERC Starting Grant (STRIPES, 101165317, PI: D.~Tzionas).}

\noindent
{\small
\textbf{Disclosure:} 
DT has received a research gift fund from Google.
For MJB see \href{https://files.is.tue.mpg.de/black/CoI\_CVPR\_2025.txt}{https://files.is.tue.mpg.de/black/CoI\_CVPR\_2025.txt}
}

{
    \small
    \bibliographystyle{config/ieeenat_fullname}
    \bibliography{config/BIB.bib}
}

\clearpage
\maketitlesupplementary

\renewcommand{\thesection}{S.\arabic{section}}
\renewcommand{\thefigure}{S.\arabic{figure}}
\renewcommand{\thetable}{S.\arabic{table}}
\renewcommand{\theequation}{S.\arabic{equation}}

\setcounter{section}{0}
\setcounter{figure}{0}
\setcounter{table}{0}
\setcounter{equation}{0}

\section{Human Contact Prediction}

\subsection{Evaluation on the 3DIR Dataset}
We evaluate our method against several state-of-the-art approaches for human contact prediction on the 3DIR dataset~\cite{yang2023piad} as shown in~\cref{tab:binary_human_lemon}. 
Our method outperforms methods that are trained on 3D training data for only humans, while it is on par with methods that use 3D data for both humans and objects.
Moreover, by eliminating the requirement for paired human-object contact training data, our method can be trained on more categories than prior work, as unpaired datasets are more varied.
This makes our method more practical for real-world applications.

\subsection{Contact Estimation Across Body Parts}

We extend our binary contact estimation's evaluation to measure our method's performance across different human body parts to ensure it captures nuanced interactions effectively.
As shown in~\cref{tab:body_parts_metrics}, our method (\nameMethod) significantly outperforms DECO~\cite{tripathi2023deco} across all body parts, including the head, torso, hands, arms, feet, and legs. 
The results demonstrate our method excels at detecting contacts across diverse body parts, making it well-suited for real-world scenarios.

\subsection{Semantic Human Contact per Object Class}
We evaluate our method's performance on ``semantic human contact'' prediction across a diverse set of object categories from the \DAMON dataset, as shown in~\cref{tab:semantic_human_contact_object}.
Results for high-level categories are presented in the main paper.
We compare our method against ``Semantic-\DECO'', which is our extension of the existing DECO~\cite{tripathi2023deco} model for this new task.
Our method significantly outperforms Semantic-\DECO in terms of F1-score for all categories.
It also demonstrates strong performance across a wide range of object categories, from large objects like furniture (couch: 62.1\% F1, chair: 70.3\% F1) to small objects for sports (baseball glove: 93.6\% F1, tennis racket: 82.3\% F1).

\section{Ablation Studies}

We conduct extensive ablation studies to evaluate the contribution of \nameMethod's main components, including the influence of the \VLM and prompt design choices. The results are summarized in \cref{tab:ablations_merged}, where we use alphabet numbering to refer to each variant for clarity. Below, we discuss the key findings from these experiments.

\qheading{Mask Resolution and \MVLoc Components}
Increasing the mask resolution from $512 \times 512$ (variant (a)) to $1024 \times 1024$ (variant (b)) yields a significant improvement of $4.9\%$ in F1 score, highlighting the importance of fine-grained spatial information for contact detection. For the \MVLoc feature embedding, using our \FeatLift network (variant (e)) outperforms simply concatenating camera parameters (variant (d)) by $3.7\%$, demonstrating its effectiveness in incorporating viewpoint information. Removing camera parameters entirely (variant (c)) further degrades performance, emphasizing their role in the pipeline. However, the performance significantly drops when we replace \MVLoc with a 2 layer MLP (variant (f)).

\begin{table}
   \centering
   \scriptsize
   \resizebox{\columnwidth}{!}{
       \begin{tabular}{l|c|c|cccc}
           \toprule
           Method & \multicolumn{2}{c|}{3D Supervision} & F1 & Prec.  & Rec.  & Geo. \\
            & Human & Obj. & (\%) $\uparrow$ & (\%) $\uparrow$ & (\%) $\uparrow$ & (cm) $\downarrow$ \\
           \midrule           
           BSTRO~\cite{RICH}            & \cmark & \xmark & 55.0 & 57.0 & 58.0 & 28.58 \\
           DECO~\cite{tripathi2023deco} & \cmark & \xmark & 69.0 & 70.0 & 72.0 & 15.25 \\
           LEMON-P~\cite{yang2023lemon} & \cmark & \cmark & 77.0 & 76.0 & 81.0 & 9.02 \\
           LEMON-D~\cite{yang2023lemon} & \cmark & \cmark & 78.0 & 78.0 & \textbf{82.0} & 7.55 \\
           \midrule
           \textbf{\nameMethod}                  & \cmark & \xmark & \textbf{78.4} & \textbf{82.5} & 76.3 & \textbf{6.73} \\
           \bottomrule
       \end{tabular}
   }
   \caption{
       Evaluation for ``Binary Human Contact'' prediction on the 3DIR dataset~\cite{yang2023lemon}.
       Note that \LEMON is trained with paired human-object contact data from 3DIR dataset. 
       Instead, for this task, \nameMethod is only trained with human contact data from the same dataset. 
   }
   \label{tab:binary_human_lemon}
\end{table}

\begin{table}
    \centering
    \scriptsize
    \resizebox{0.99 \columnwidth}{!}{
        \begin{tabular}{l|ccccccc}
            \toprule
            Method & Head & Torso & Hips & Hands & Arms & Feet & Legs \\
            \midrule
            DECO~\cite{tripathi2023deco} & 20.0 & 46.1 & 66.6 & 74.3 & 22.2 & 94.4 & 66.6 \\
            \textbf{Ours} & \textbf{56.0} & \textbf{87.2} & \textbf{95.7} & \textbf{93.5} & \textbf{71.5} & \textbf{96.9} & \textbf{68.3} \\
            \bottomrule
        \end{tabular}
    }
    \vspace{-1.0 em}
    \caption{F1 scores for human contact estimation \wrt body parts}
    \label{tab:body_parts_metrics}
    \vspace{-1.0 em}
\end{table}

\qheading{Loss Functions}
Using only the valid mask regions for training (variant (h)) improves performance by $3.3\%$ compared to using the whole mask (variant (g)). The addition of our 3D contact loss (variant (i)) further boosts the F1 score by $3\%$, underscoring the importance of explicitly modeling 3D contact cues during training.

\qheading{Data and \VLM Influence}
The choice of training data significantly impacts performance. Using only 3D contact datasets (variant (j)) results in a relatively low F1 score of $65.9\%$. Adding contact parts in text form (variant (k)) improves performance by $8.9\%$, while further incorporating HOI-VQA data (variant (l)) achieves the best results. This demonstrates the value of leveraging textual and contextual cues for contact localization.

The \VLM plays a critical role in the pipeline. Removing the \VLM entirely (variant (m)) drastically reduces performance, while using a \VLM with only image input (variant (n)) serves as a strong baseline. Fine-tuning the \VLM (variant (b)) is crucial, as the non-fine-tuned version (variant (o)) shows a significant drop in performance. Interestingly, reducing the \VLM size from 13B to 7B parameters (variant (p)) has minimal impact, suggesting that the model can maintain strong performance even with fewer parameters.

\begin{table}
    \vspace{-0.5 em}
    \centering
    \resizebox{\columnwidth}{!}{
        \begin{tabular}{l|c|cccc|cccc}
            \toprule
            Object & \# & \multicolumn{4}{c|}{Semantic-\DECO~\cite{tripathi2023deco}} & \multicolumn{4}{c}{\textbf{\nameMethod (Ours)}} \\
            \cmidrule{3-6} \cmidrule{7-10}
             &  & F1 & Prec. & Rec. & Geo. & F1 & Prec. & Rec. & Geo. \\
            Categories  & & (\%) $\uparrow$ & (\%) $\uparrow$ & (\%) $\uparrow$ & (cm) $\downarrow$ & 
            (\%) $\uparrow$ & (\%) $\uparrow$ & (\%) $\uparrow$ & (cm) $\downarrow$ \\
            \midrule
       Skateboard & 85 & 30.3 & 19.3 & \textbf{91.3} & 99.95 & \textbf{71.5} & \textbf{67.0} & 83.5 & \textbf{0.90} \\
           Surfboard & 70 & 23.1 & 14.2 & \textbf{98.4} & 101.22 & \textbf{79.7} & \textbf{76.3} & 78.9 & \textbf{0.80} \\
           Snowboard & 49 & 38.2 & 25.7 & \textbf{92.2} & 108.29 & \textbf{84.2} & \textbf{83.1} & 84.0 & \textbf{0.20} \\
           T. Racket & 45 & 57.0 & 42.0 & \textbf{99.6} & 64.25 & \textbf{82.3} & \textbf{80.8} & 86.3 & \textbf{0.20} \\
           Cell phone & 43 & 42.4 & 27.8 & \textbf{99.6} & 51.73 & \textbf{70.6} & \textbf{73.1} & 74.3 & \textbf{7.00} \\
           Couch & 38 & 31.4 & 19.7 & \textbf{89.2} & 17.07 & \textbf{62.1} & \textbf{62.5} & 60.5 & \textbf{2.10} \\
           Bicycle & 37 & 62.1 & 48.0 & \textbf{98.1} & 29.89 & \textbf{81.5} & \textbf{84.4} & 81.9 & \textbf{2.50} \\
           Chair & 36 & 23.2 & 14.6 & \textbf{87.1} & 36.05 & \textbf{70.3} & \textbf{73.6} & 68.8 & \textbf{1.60} \\
           Bench & 35 & 19.0 & 11.2 & \textbf{92.1} & 29.51 & \textbf{63.0} & \textbf{70.7} & 64.4 & \textbf{4.00} \\
           Motorcycle & 33 & 60.4 & 45.5 & \textbf{99.1} & 19.24 & \textbf{76.6} & \textbf{78.6} & 77.7 & \textbf{0.90} \\
           Book & 27 & 48.0 & 33.8 & \textbf{99.7} & 53.59 & \textbf{74.1} & \textbf{75.2} & 80.1 & \textbf{1.10} \\
           Skis & 25 & 36.5 & 25.0 & \textbf{93.4} & 104.07 & \textbf{83.0} & \textbf{81.4} & 83.7 & \textbf{0.80} \\
           Bed & 24 & 29.1 & 19.1 & \textbf{82.9} & 20.71 & \textbf{54.0} & \textbf{56.7} & 48.8 & \textbf{2.70} \\
           Laptop & 24 & 36.9 & 24.9 & \textbf{94.4} & 45.73 & \textbf{54.0} & \textbf{54.0} & 68.6 & \textbf{4.70} \\
           Backpack & 24 & 37.2 & 24.3 & \textbf{87.2} & 12.10 & \textbf{59.2} & \textbf{71.1} & 54.8 & \textbf{3.50} \\
           Umbrella & 23 & 51.5 & 36.1 & \textbf{99.2} & 67.20 & \textbf{82.3} & \textbf{83.7} & 86.4 & \textbf{1.00} \\
           Knife & 19 & 63.3 & 54.0 & 84.4 & 31.55 & \textbf{77.0} & \textbf{74.9} & \textbf{86.6} & \textbf{0.10} \\
           Frisbee & 15 & 33.9 & 22.0 & \textbf{99.4} & 69.43 & \textbf{68.7} & \textbf{71.5} & 84.5 & \textbf{1.00} \\
           D. Table & 11 & 19.6 & 14.1 & \textbf{67.1} & 42.56 & \textbf{35.2} & \textbf{44.9} & 63.4 & \textbf{6.60} \\
           B. Glove & 10 & 71.4 & 63.3 & 81.9 & 41.58 & \textbf{93.6} & \textbf{98.6} & \textbf{89.1} & \textbf{0.10} \\
           Remote & 10 & 0.2 & 1.0 & 0.1 & 82.16 & \textbf{70.6} & \textbf{77.4} & \textbf{82.7} & \textbf{0.50} \\
           Banana & 10 & 6.1 & 7.1 & 6.4 & 67.19 & \textbf{76.6} & \textbf{74.3} & \textbf{81.7} & \textbf{2.80} \\
           Kite & 9 & 65.3 & 51.8 & \textbf{95.9} & 50.50 & \textbf{85.4} & \textbf{86.0} & 85.4 & \textbf{0.30} \\
           Toothbrush & 8 & 2.9 & 4.7 & 2.1 & 56.38 & \textbf{77.3} & \textbf{82.6} & \textbf{74.8} & \textbf{5.40} \\
           Boat & 8 & 33.5 & 23.9 & \textbf{83.7} & 46.24 & \textbf{71.3} & \textbf{75.3} & 63.1 & \textbf{1.40} \\
           Sports ball & 8 & 36.0 & 34.1 & 39.4 & 60.54 & \textbf{64.4} & \textbf{74.0} & \textbf{83.8} & \textbf{5.30} \\
           B. Bat & 8 & 36.7 & 60.8 & 27.2 & 26.00 & \textbf{82.8} & \textbf{81.2} & \textbf{87.8} & \textbf{1.60} \\
           Apple & 7 & 6.3 & 17.4 & 3.9 & 45.69 & \textbf{69.3} & \textbf{62.9} & \textbf{77.7} & \textbf{4.20} \\
           Handbag & 7 & 12.1 & 7.0 & \textbf{46.2} & 26.61 & \textbf{31.8} & \textbf{27.1} & 40.4 & \textbf{4.10} \\
           Tie & 6 & 39.8 & 28.1 & \textbf{87.2} & \textbf{7.24} & \textbf{49.6} & \textbf{32.8} & 60.8 & 7.60 \\
           Suitcase & 6 & 26.7 & 24.0 & 30.7 & 87.44 & \textbf{79.2} & \textbf{65.9} & \textbf{83.4} & \textbf{0.80} \\
           Wine glass & 5 & 5.5 & 8.4 & 5.0 & 70.32 & \textbf{66.4} & \textbf{68.5} & \textbf{69.4} & \textbf{4.40} \\
           Spoon & 5 & 61.1 & 48.5 & \textbf{89.9} & 15.35 & \textbf{67.5} & \textbf{62.8} & 78.5 & \textbf{5.50} \\
           Fork & 5 & 1.5 & 1.6 & 1.3 & 75.47 & \textbf{64.9} & \textbf{66.2} & \textbf{76.5} & \textbf{2.20} \\
           Keyboard & 5 & 3.2 & 6.2 & 3.1 & 70.41 & \textbf{60.8} & \textbf{69.1} & \textbf{74.0} & \textbf{0.50} \\
           Teddy bear & 5 & 17.5 & 15.7 & 45.0 & 24.70 & \textbf{43.8} & \textbf{61.6} & \textbf{68.8} & \textbf{11.60} \\
           Clock & 4 & 23.3 & 14.8 & 58.1 & 46.42 & \textbf{37.1} & \textbf{68.9} & \textbf{75.0} & \textbf{3.30} \\
           Cake & 4 & 0.0 & 0.0 & 0.0 & 83.99 & \textbf{52.4} & \textbf{41.9} & \textbf{82.2} & \textbf{10.60} \\
           Scissors & 4 & 0.2 & 0.2 & 0.2 & 87.88 & \textbf{28.7} & \textbf{21.4} & \textbf{73.1} & \textbf{40.10} \\
           Cup & 4 & 7.2 & 11.2 & 5.4 & 69.03 & \textbf{68.6} & \textbf{71.4} & \textbf{76.2} & \textbf{1.70} \\
           Car & 4 & 0.0 & 0.0 & 0.0 & 49.13 & \textbf{66.7} & \textbf{67.7} & \textbf{73.3} & \textbf{5.30} \\
           Pizza & 4 & 19.4 & 19.0 & 35.1 & 46.43 & \textbf{44.3} & \textbf{44.1} & \textbf{71.4} & \textbf{29.20} \\
           Carrot & 3 & 0.0 & 0.0 & 0.0 & 90.22 & \textbf{59.7} & \textbf{62.4} & \textbf{77.6} & \textbf{0.20} \\
           Truck & 3 & 0.0 & 0.0 & 0.0 & 61.65 & \textbf{81.2} & \textbf{84.9} & \textbf{77.5} & \textbf{3.10} \\
           Bottle & 3 & 0.0 & 0.0 & 0.0 & 91.14 & \textbf{59.2} & \textbf{55.1} & \textbf{81.2} & \textbf{0.10} \\
           Airplane & 2 & 0.0 & 0.0 & 0.0 & 87.52 & \textbf{76.4} & \textbf{69.3} & \textbf{85.2} & \textbf{3.60} \\
           Toilet & 2 & 0.0 & 0.0 & 0.0 & 86.55 & \textbf{32.5} & \textbf{35.7} & \textbf{71.1} & \textbf{3.30} \\
           Hot dog & 2 & 7.0 & 23.0 & 4.1 & 46.32 & \textbf{81.3} & \textbf{84.0} & \textbf{78.9} & \textbf{4.10} \\
           Donut & 2 & 19.6 & 30.7 & 14.8 & 42.47 & \textbf{73.6} & \textbf{90.1} & \textbf{65.6} & \textbf{12.00} \\
           Mouse & 1 & 0.0 & 0.0 & 0.0 & 82.03 & \textbf{40.7} & \textbf{27.0} & \textbf{82.9} & \textbf{0.10} \\
           Vase & 1 & 0.0 & 0.0 & 0.0 & 91.96 & \textbf{68.5} & \textbf{59.3} & \textbf{81.0} & \textbf{0.20} \\
            F. Hydrant & 1 & 0.0 & 0.0 & 0.0 & 88.18 & \textbf{85.5} & \textbf{82.7} & \textbf{88.5} & \textbf{0.00} \\
            \bottomrule
        \end{tabular}
    }
    \caption{Evaluation for “Semantic Human Contact” prediction on
    the DAMON~\cite{tripathi2023deco} dataset for different object categories in the test set. The number of samples for each category is shown in the second column.
    ``Semantic-\DECO'' is our extension of the existing \DECO~\cite{tripathi2023deco} model for this new task.
    Zero metrics indicate no correct     predictions for the class.
    }
    \label{tab:semantic_human_contact_object}
\end{table}

\begin{table}
    \vspace{-0.5 em}
    \centering
    \scriptsize
    \resizebox{1.00 \columnwidth}{!}{
        \begin{tabular}{l|l|c|c|c}
            \toprule
            & Variants & F1  & Prec.  & Rec. \\
            &          & (\%) $\uparrow$ & (\%) $\uparrow$ & (\%) $\uparrow$ \\
            \midrule
            \multirow{2}{*}{Masks}
                    & (a) Size $512\times512$       & 70.7 & 70.1 & 71.4     \\
                    & (b) Size $1024\times1024$     & 75.6 & 75.2 & 76.0     \\
            \midrule
            \multirow{4}{*}{\MVLoc}
                    & (c) No CamParams              & 69.4 & 68.0 & 71.1     \\
                    & (d) Concat CamParams          & 71.9 & 72.0 & 71.8     \\
                    & (e) \FeatLift ($\featLiftNN$) & 75.6 & 75.2 & 76.0    \\
                    & (f) No \MVLoc                 & 62.3 & 60.8 & 63.9    \\
            \midrule
            \multirow{3}{*}{Losses}
                    & (g) Whole Mask                & 69.3 & 68.7 & 70.0    \\
                    & (h) Valid Mask                & 72.6 & 71.2 & 74.0    \\
                    & (i) + 3D Contact Loss         & 75.6 & 75.2 & 76.0    \\
            \midrule
            \multirow{3}{*}{Data}
                    & (j) 3D Contact Datasets       & 65.9 & 64.8 & 67.0    \\
                    & (k) + Contact Parts (text)    & 74.8 & 74.5 & 75.1    \\
                    & (l) + HOI-VQA                 & 75.6 & 75.2 & 76.0    \\
            \midrule
            \multirow{4}{*}{\VLM}
                    & (m) No \VLM                   & 32.3 & 30.8 & 43.0    \\
                    & (n) \VLM-13B-Img              & 67.2 & 68.5 & 66.0    \\
                    & (o) \VLM-13B-NoFT             & 64.8 & 65.3 & 64.2    \\
                    & (p) \VLM-7B                   & 73.3 & 76.8 & 73.5    \\
            \midrule
            \multirow{3}{*}{Prompt}
                    & (q) Contact parts (fine)      & 74.8 & 74.5 & 75.1    \\
                    & (r) Contact parts (coarse)    & 68.4 & 69.0 & 67.8    \\
                    & (s) No object name            & 71.5 & 72.1 & 70.9    \\
            \bottomrule
        \end{tabular} 
    }
    \vspace{-0.5 em}
    \caption{
        \CR{
        Ablation study for the effect of different \nameMethod components. We evaluate for ``Binary Human Contact'' prediction on the \DAMON dataset~\cite{tripathi2023deco}.
        }
    }
    \label{tab:ablations_merged}
\end{table}

\qheading{Prompt Design}
The design of the text prompt significantly influences the results. Using fine-grained contact parts (variant (q)) outperforms a coarse segmentation (variant (r)) by $6.4\%$ in F1 score, indicating that finer granularity in body part labeling is beneficial. Removing the object name from the prompt (variant (s)) also degrades accuracy, highlighting the importance of explicit object context in guiding the \VLM's predictions.

Our ablation studies demonstrate the importance of fine-grained spatial information, effective feature embedding, 3D contact modeling, and well-designed prompts in achieving robust contact localization. The \VLM's role, particularly its fine-tuning and input modalities, is also critical to the overall performance.

\section{Impact of RLL}

``\nameRepresentationLONG'' (\nameRepresentation) is central to our method. 
Traditional approaches, like DECO~\cite{tripathi2023deco} and RICH~\cite{RICH}, rely on fully supervised learning with limited 3D GT data to predict 3D contacts.
While effective for scenarios encountered during training, these methods fail to generalize to \inthewild cases.
To address this limitation, we leverage the broad visual knowledge of \VLM to learn from the limited data. 
However, effectively utilizing \VLM requires reformulating our 3D problem into a 2D representation, making it compatible with \VLM, which we achieve through \nameRepresentation. 
As demonstrated in the main paper, by using RLL with \VLM, we surpass the \sota method for human contact estimation while training on only 5\% of the data.

\section{Implementation Details}

\subsection{Architecture}

\nameMethod has two major blocks; a reasoning module, $\vlmNN$, based on \LLAVA-v1~\cite{liu2023llava} and a novel multi-view localization model, \MVLoc, based on \SAM~\cite{sam}.
\MVLoc has 2 components; a shared encoder, $\imageEncoder$ and two separate 2D contact decoders, $\HmaskDecoder$ and $\OmaskDecoder$, for humans and objects respectively. 
$\imageEncoder$, $\HmaskDecoder$, and $\OmaskDecoder$ have the same architecture as \SAM.

Given an RGB image, $\image$, and prompt text, $\inputPrompt$, the \VLM produces contact tokens, \humanToken and \objectToken, for humans and objects, respectively. 
To aid \MVLoc in localizing contact, we extract the last-layer embeddings of the \VLM corresponding to these tokens and pass them through a projection layer, $\projLayerNN$. 
The latter, $\projLayerNN$, is a multi-layer perceptron with 2 layers each of size $256$ and a ReLU activation.

\subsection{Training}
Before the start of training, we render multiple views of the human mesh and object point cloud. We also compute the ground-truth contact mask.

\subsubsection{Human Mesh Rendering}
The human mesh rendering pipeline uses a comprehensive multi-view approach using the \smplH~\cite{romero2022mano_smplh} parametric body model. 
We initialize the model in a neutral shape, positioning the body in a Vitruvian pose. This specific pose ensures optimal visibility of potential contact surfaces.
We use \mbox{PyTorch3D} for rendering. 
We select 4 camera viewpoints to capture the complete body geometry: top-front (elevation 45°, azimuth 315°), top-back (45°, 135°), bottom-front (315°, 315°), and bottom-back (315°, 135°). 
Each viewpoint is positioned at a distance of 2 units from the subject with slight horizontal translations to optimize coverage. 
We use a FoV-Perspective projection model rendered at 1024×1024 resolution, with ``blur-radius'' and ``faces-per-pixel'' settings set as $0.0$ and $1$, respectively.
For realistic appearance, we use point lights positioned at [0, 0, ±3] coordinates relative to the mesh. 
The lighting settings such as ``ambient'', ``diffuse'', and ``specular'' are set at $0.5$, $0.3$, $0.2$, respectively. 
This creates a balanced illumination that highlights surface details. 
Surface normals are computed per vertex and are used as vertex colors.

Crucially, \nameMethod maintains precise correspondence between 2D rendered pixels and 3D mesh vertices. For each rendered view, it generates:
(1) A pixel-to-vertex mapping matrix storing the indices of mesh vertices visible at each pixel.
(2) Barycentric coordinates for accurate interpolation within mesh faces.
(3) Binary contact masks for regions with at least three neighboring vertices in contact.

This comprehensive multi-view representation, combined with precise pixel-to-vertex correspondences, enables accurate lifting of 2D contact predictions back to the 3D mesh space.
Our model processes each view as separate channels in a $B\times V \times 3 \times H \times W$ tensor shape during training, where B is the batch size and V is the number of views.

\subsubsection{Object Point Cloud Rendering}
The object rendering pipeline uses point clouds to capture object affordances in multiple views. 
The point cloud pre-processing begins with normalization, where each object is centered at its geometric centroid and scaled to fit within a unit sphere, ensuring consistent scale across different objects. 
Since the point clouds do not have color, we use the NOCS representation for coloring, namely 
for every point we assign a color derived from its normalized spatial NOCS coordinates (scaled to [0.1, 0.9] for better contrast).

Our rendering pipeline uses \mbox{PyTorch3D} with four viewpoints: front-left (elevation 45°, azimuth 315°), front-right (45°, 45°), back-left (330°, 135°), and back-right (330°, 225°). 
Each view is rendered at 1024×1024 resolution using a \mbox{FoVPerspective} camera positioned at a distance of 2 units from the object center.
We use a fixed point cloud radius of $0.05$.
For the rasterization settings: we use 10 points per pixel and 50,000 points per bin to handle dense point clouds effectively. 
An alpha compositor is used for the final rendering. 
For affordance heatmaps, we generate a rendered view with continuous values, $[0,1]$, representing the affordance likelihood. 
For each view, we create a pixel-to-point mapping for lifting 2D affordance heatmaps to 3D affordance points.

\subsection{Additional Text Data for Training}

\subsubsection{Data from GPT4o}
To enhance our model's understanding of human-object interactions (\HOI), we build a comprehensive Visual Question-Answering (\VQA) data generation pipeline using \mbox{GPT-4V} (\mbox{GPT4o}). 
The pipeline processes images from three datasets, namely \DAMON~\cite{tripathi2023deco}, \LEMON~\cite{yang2023lemon}, and \PIAD~\cite{yang2023piad}, generating structured textual descriptions that capture multiple aspects of \HOI.

For each image, we query \mbox{GPT-4V} to describe five key aspects: 
(1) the human's visual appearance including clothing and distinctive features, 
(2) specific body parts in contact with the object, 
(3) the nature of the interaction, 
(4) the object's physical characteristics, and 
(5) the specific parts of the object in contact with the human. 
To ensure efficient processing while maintaining visual fidelity, images are resized to 256$\times$256 pixels.

These generated \VQA data enrich the training signal with detailed descriptions of interactions.
This additional supervision helps our model develop a more nuanced understanding of the relationship between visual features and contact regions, ultimately contributing to improved performance in contact prediction tasks.
We format the collected data as JSON files to seamlessly integrate these with our \VLM training pipeline, allowing the model to leverage these rich textual descriptions during the learning process.

\begin{figure}
    \centering%
    \includegraphics[trim=000mm 000mm 000mm 000mm, clip=true, width=0.99 \columnwidth]{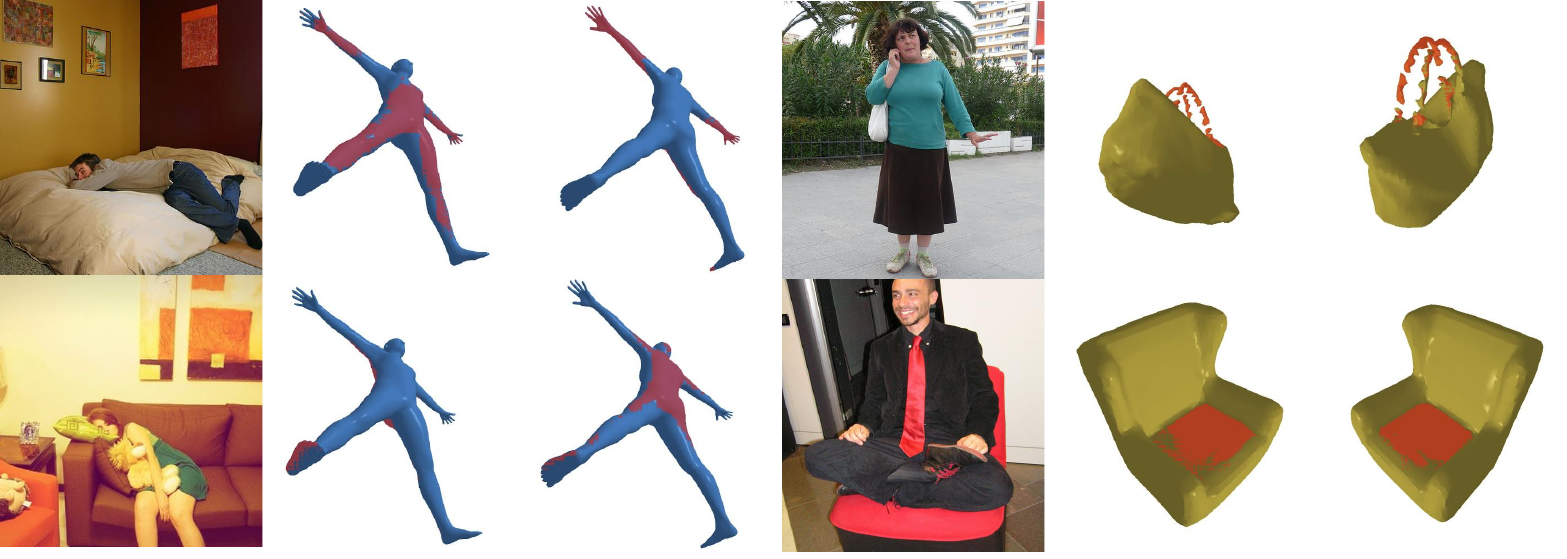}
    \caption{
        \qheading{Contact Estimation Failure Cases}
        Our method struggles with unusual human poses (left). %
        For objects (right), training on affordances rather than actual contacts can sometimes lead to ambiguous contact predictions, especially for large objects like chairs.
        However, no dataset exists for 3D object contacts for \inthewild images. 
    }
    \label{fig:failure_cases}
\end{figure}

\begin{figure}
    \centering%
    \includegraphics[width=0.99 \columnwidth]{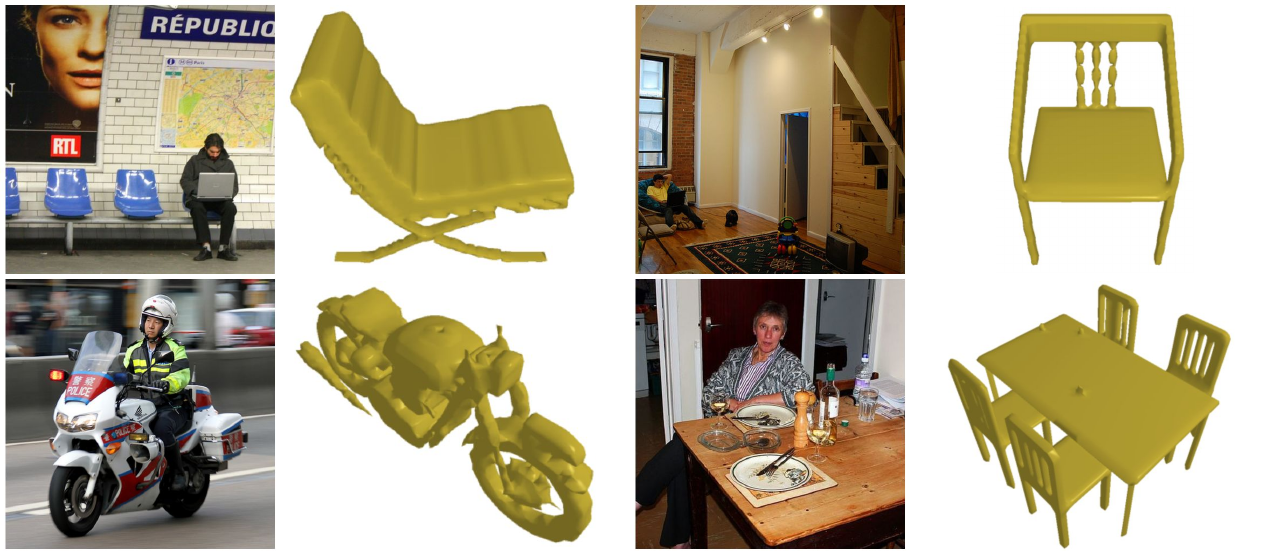}
    \caption{
        \qheading{Object Retrieval Failure Cases}
        The retrieved object meshes (right) differ notably from the actual objects in the input images (left), especially in cases of significant occlusion, atypical object instances, or limited database coverage. Despite these inaccuracies, the retrieval consistently selects objects within the correct semantic category.
    }
    \label{fig:lookup_failure}
    \vspace{0.5em}
\end{figure}

\subsubsection{Converting 3D contact vertices to text}
To establish a precise mapping between 3D contact vertices and natural-language descriptions, we leverage the \smpl body model's semantic segmentation. 
The body is divided into 15 semantically meaningful parts including the torso, head, hands, feet, arms, legs, thigh and forearm. 
For training our \VLM, we employ a diverse set of natural-language prompts that query about body part contacts with objects. 
This structured approach creates a strong bridge between geometric contact information and natural-language understanding, enabling the model to learn the relationship between visual features, contact regions, and their semantic descriptions.

\section{Failure Cases}
Despite the overall strong performance, our method has certain limitations. 
For human contact prediction, our method occasionally struggles with unusual or ambiguous poses that deviate significantly from common interaction patterns.
For example, in~\cref{fig:failure_cases} the person is sleeping in an unusual pose on the bed.

Regarding objects, 
our method faces challenges inherent to the training paradigm. 
Since  there exists no dataset of \inthewild images with ground-truth 3D contact annotations for objects, we train on affordance data, which represents likelihood of contact rather than actual contact points. 
However, 
the distinction between actual contacts and affordances can be ambiguous, particularly 
for large objects like chairs, as shown in~\cref{fig:failure_cases}.

In highly occluded or visually ambiguous scenarios, our approach can face challenges due to object lookup failure.
The object lookup is also limited by the richness and diversity of underlying object database.
However, since our method performs retrieval within predefined object categories, it consistently retrieves an object instance belonging to the correct semantic category, even if exact geometric matches are not always guaranteed.
We provide qualitative examples highlighting these limitations in~\cref{fig:lookup_failure}.

\section{Qualitative Results}
We present qualitative results for our \nameMethod method for three different tasks.
First, in~\cref{fig:qualitative-object-affordance} we show the object affordance prediction results, where our method more accurately identifies plausible contact regions on objects compared to the state-of-the-art \mbox{IAGNet} method.
Second, we show ``semantic human contact'' prediction results in~\cref{fig:qualitative-semantic-supplementary}, where our method successfully identifies contact regions on human bodies specific to different object categories, even in complex scenarios.
Finally, in~\cref{fig:qualitative-3D-HOI-supp}, we demonstrate 3D HOI reconstruction from in-the-wild images, where we leverage the \emph{inferred} contacts on \emph{both} human bodies and objects to generate physically plausible 3D reconstructions; this is done for the first time for \inthewild images.

\section{Future Work}
Our approach follows a two-stage process for 3D \HOI: 
it first predicts human and object contacts, and then uses the inferred contacts in optimization for joint 3D reconstruction. 
In the future, 
we will explore learning to perform both 3D contact prediction and 3D reconstruction in an end-to-end fashion. 
This could lead to more coherent predictions by learning and exploiting direct relationships between contact points and physical constraints. 

Moreover, currently our approach learns on disjoint image datasets of body contacts and object affordances. 
In the future, we will also exploit recent datasets of images paired with contact annotations for both bodies and objects \cite{cseke2025pico}.

\clearpage

\begin{figure*}
    \vspace{-0.8 em}
    \centering
    \includegraphics[width=0.92\textwidth]{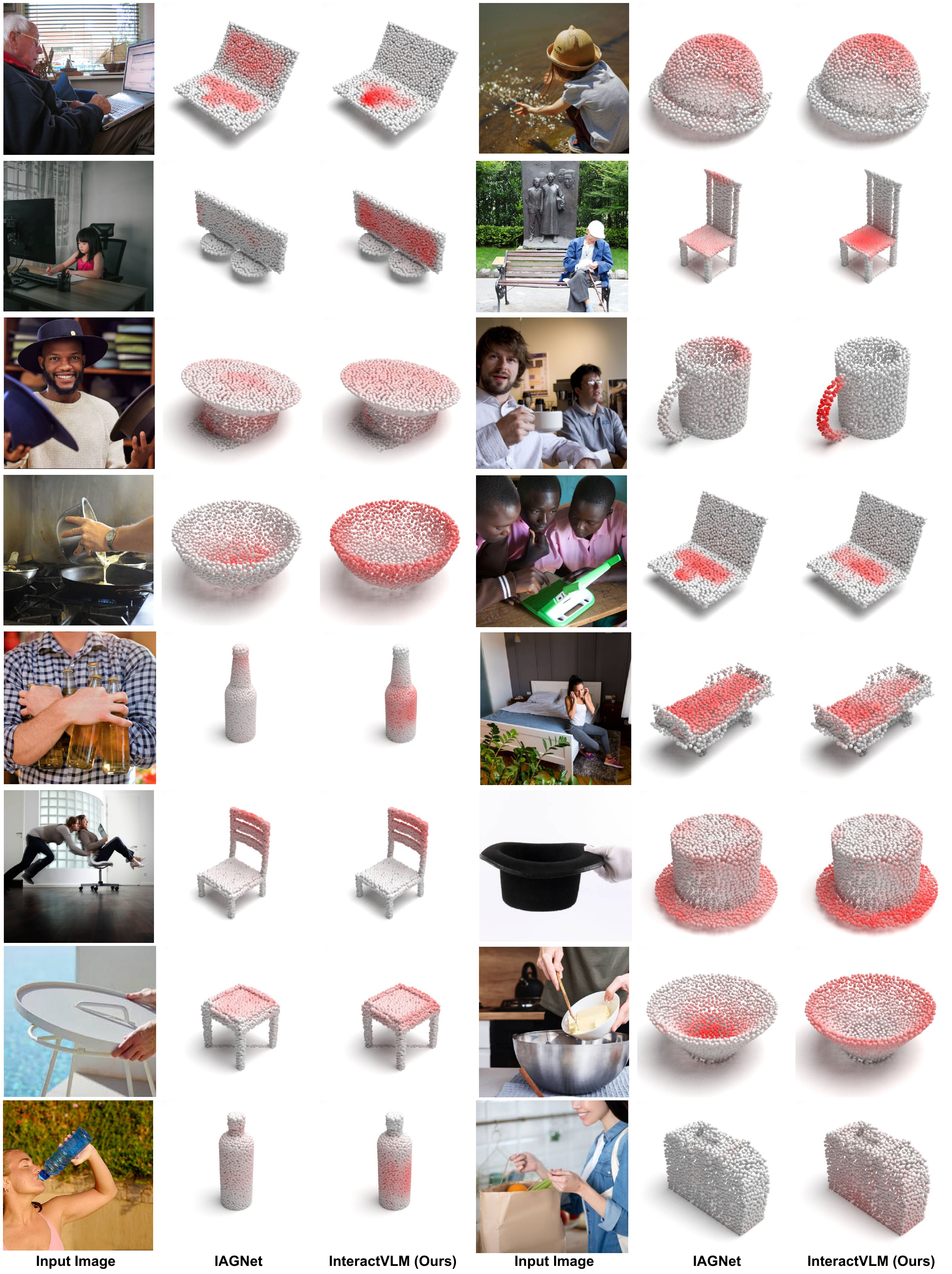}
    \vspace{-0.5em}
    \caption{
                \qheading{Object Affordance Prediction} 
                Here we compare our \nameMethod method trained for object affordance prediction on \PIAD~\cite{yang2023piad} dataset with the \sota \mbox{IAGNet} method.
                We train for affordance detection because there exists no dataset of in-the-wild images paired with ground-truth 3D contacts for objects. 
                Note that given an image of a person performing an action like ``sit'' or ``grasp'', the affordance prediction task estimates ``contact possibilities''
                on the object.
            }
    \label{fig:qualitative-object-affordance}
\end{figure*}

\begin{figure*}
    \vspace{-0.8 em}
    \centering
    \includegraphics[width=0.93\textwidth]{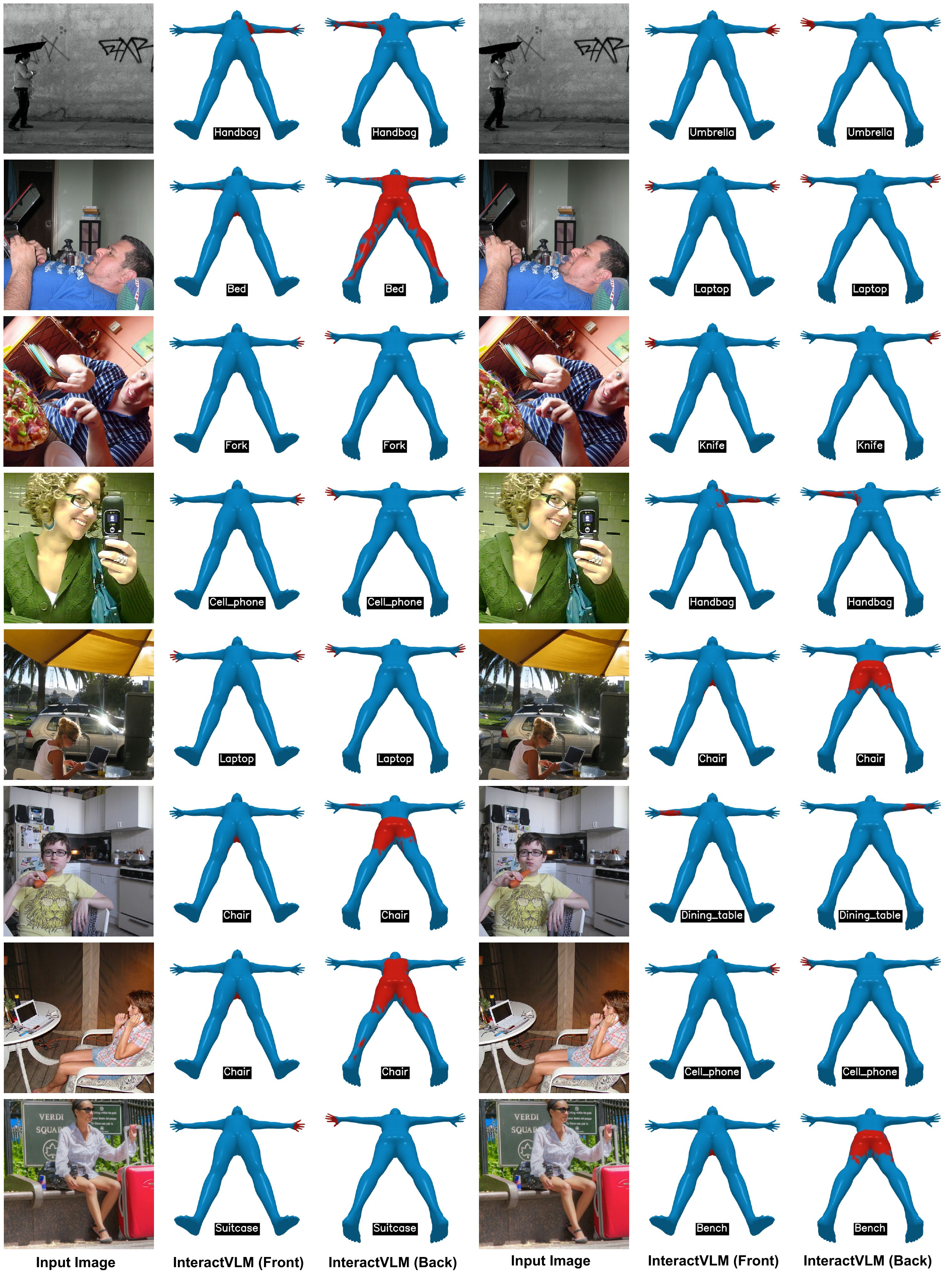}
    \vspace{-0.5em}
    \caption{
                \qheading{Semantic Human Contact estimation}
                Here we show results for ``semantic human contact'' estimation from \inthewild images. 
                Each row shows a person in contact with multiple objects. 
                Note how \nameMethod estimates contact on bodies that is specific to the object.
    }
    \label{fig:qualitative-semantic-supplementary}
\end{figure*}

\begin{figure*}
    \vspace{-0.8 em}
    \centering
    \includegraphics[width=0.95\textwidth]{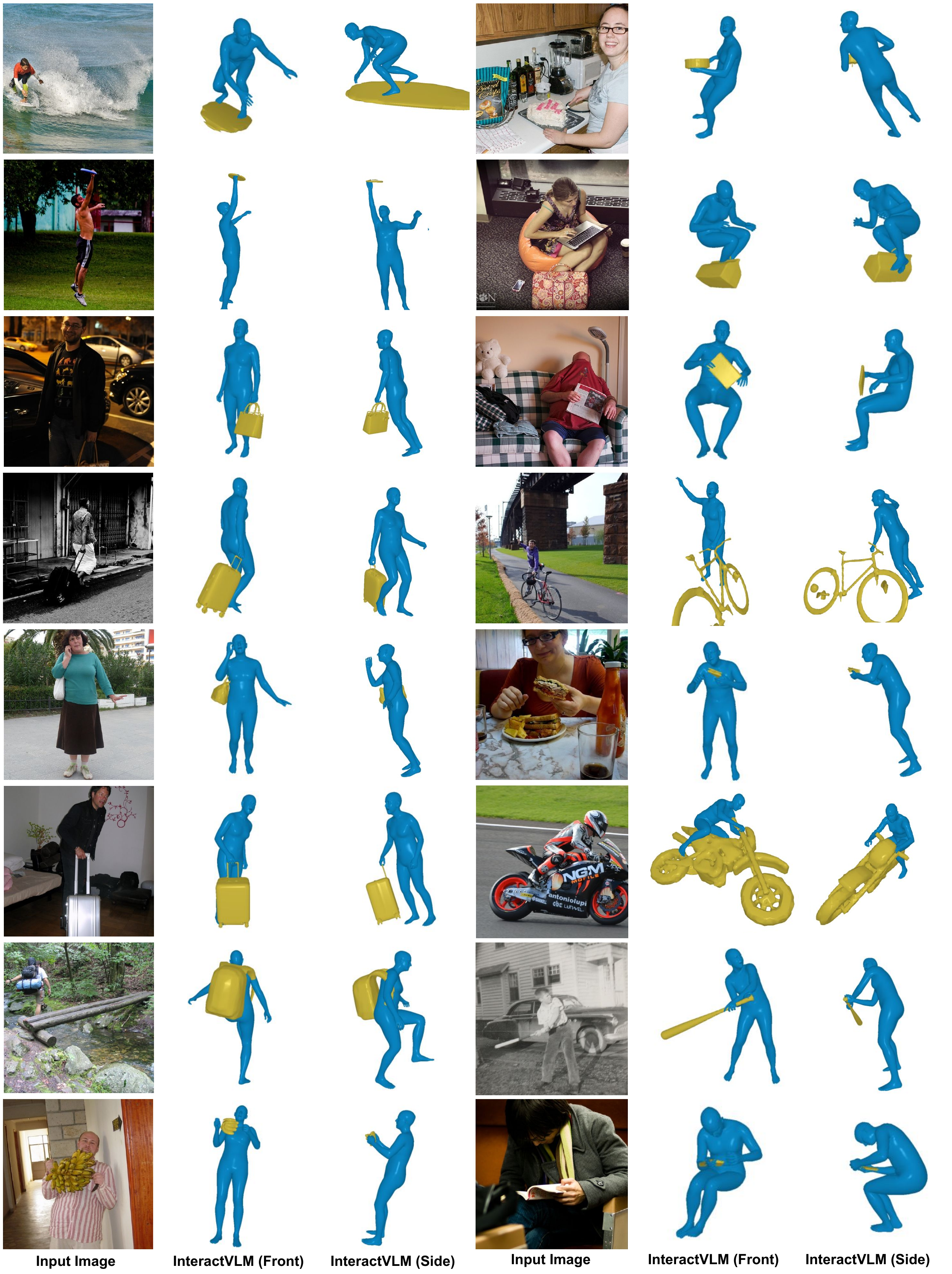}
    \vspace{-0.5em}
    \caption{
                \qheading{3D \HOI reconstruction}
                Here we show results of our \nameMethod method for 3D \HOI reconstruction from \inthewild images. 
                We use the \nameMethod's inferred contacts on both bodies and objects for joint 3D reconstruction. 
    }
    \label{fig:qualitative-3D-HOI-supp}
\end{figure*}

\end{document}